%% file: main.tex
\documentclass{article}

\usepackage[preprint]{neurips_2026}

\usepackage[utf8]{inputenc} % allow utf-8 input
\usepackage[T1]{fontenc}    % use 8-bit T1 fonts
\usepackage{hyperref}       % hyperlinks
\usepackage{url}            % simple URL typesetting
\usepackage{booktabs}       % professional-quality tables
\usepackage{amsfonts}       % blackboard math symbols
\usepackage{nicefrac}       % compact symbols for 1/2, etc.
\usepackage{microtype}      % microtypography
\usepackage{xcolor}         % colors
\usepackage{graphicx}
\usepackage{amsmath}
\usepackage{amssymb}
\usepackage{amsthm}
\usepackage{mathtools}
\usepackage{booktabs}
\usepackage{multirow}
\usepackage{booktabs}
\usepackage{longtable}
\usepackage{pdflscape}
\usepackage{xcolor}
\usepackage{array}

\newtheorem{proposition}{Proposition}

\title{PA-RNet: Perturbation-Aware Residual Network for Robust Multimodal Time Series Forecasting}

\author{
    Enqiang Zhu\textsuperscript{\rm 1},
    Zhenbin Deng\textsuperscript{\rm 1},
    Shengzhi Wang\textsuperscript{\rm 2,*},
    Yi-Kun Tang\textsuperscript{\rm 2},
    Chanjuan Liu\textsuperscript{\rm 2,*}
    \\
    \textsuperscript{1}Institute of Computing Technology, Guangzhou University, Guangzhou 510006, China
    \\
    \textsuperscript{2}School of Computer Science and Technology, Dalian University of Technology, Dalian 116024, China
    \\
    \textsuperscript{*}Corresponding authors: wangshengzhi@mail.dlut.edu.cn; chanjuanliu@dlut.edu.cn
}
\begin{document}

\maketitle

\begin{abstract}
In real-world applications, multimodal time-series forecasting faces a key challenge: textual information is often useful but unreliable. Auxiliary texts may contain irrelevant, ambiguous, incomplete, or structurally corrupted content, making direct text integration prone to introducing noisy semantic signals and degrading forecasting performance. Therefore, robust multimodal forecasting requires a model that can exploit useful textual context while suppressing misleading perturbations.
To address this challenge, we propose PA-RNet, a carefully designed perturbation-aware residual network for robust multimodal time-series forecasting. Rather than directly fusing textual and numerical representations, PA-RNet first refines multimodal features in a perturbation-aware manner, preserving stable contextual information while reducing unstable or misleading signals. The refined textual representations are then aligned with temporal dynamics, enabling more reliable forecasting under noisy multimodal conditions.
Theoretically, we prove that PA-RNet is Lipschitz continuous with respect to textual embeddings and show that the proposed spectral residual correction can reduce the expected prediction error under zero-mean textual perturbations. We further conduct supplementary experiments with injected textual perturbations to examine the robustness of PA-RNet. The results across diverse domains demonstrate that PA-RNet consistently outperforms state-of-the-art baselines and maintains stable forecasting performance under both original and noise-perturbed textual conditions.\footnote{Code and Datasets are provided in the supplementary materials accompanying this paper.}
\end{abstract}

\section{Introduction}
Time series forecasting is crucial in various real-world applications where predicting future dynamics is essential. For instance, in energy management, accurate forecasting enables efficient load balancing and demand-response planning~\cite{1}; in financial markets, it assists in risk assessment and algorithmic trading strategies~\cite{2}; in healthcare monitoring, predictive models support early diagnosis and timely interventions~\cite{3}.
and in traffic systems, forecasting traffic flow patterns helps reduce congestion and enhances route planning~\cite{4}. 

In the era of data abundance, time series data are often accompanied by rich textual context, such as news articles, policy updates, or event descriptions. We refer to this combination as textual-numerical time series. Integrating qualitative context with quantitative data is essential for improved forecasting~\cite{5}, as it enables models to grasp underlying causes and temporal dynamics. For example, aligning stock prices with news can reveal factors driving market volatility~\cite{6}, while incorporating weather alerts or event notices can enhance traffic predictions~\cite{7}. This paradigm mirrors human reasoning as we combine numerical trends with contextual cues for better-informed decisions.

Existing approaches for text-numerical time series forecasting often depend on carefully curated textual inputs, which are manually crafted or selected with domain-specific knowledge or pre-processed using large language models~\cite{8,9,R13}. However, in real-world scenarios, textual information is frequently noisy, redundant, or semantically inconsistent~\cite{8}, posing significant challenges for accurate forecasting. Most current methods overlook this issue, particularly when the noise varies in intensity or arises from structural inconsistencies.

% To systematically address this issue,we investigate the impact of textual perturbations on multimodal time series forecasting performance. We introduce PA-RNet (Perturbation-Aware Reasoning Network for Multimodal Time Series Forecasting), a robust and generalizable framework designed to achieve high predictive accuracy even under significant textual noise. PA-RNet integrates a projection-based denoising module and cross-modal attention mechanism to perform perturbation-aware reasoning, effectively filtering out noisy or inconsistent textual information. Theoretically, we prove that PA-RNet satisfies Lipschitz continuity with respect to textual inputs, leading to a reduction in expected prediction error under perturbations and offering strong guarantees of robustness and generalization in real-world multimodal scenarios.
To address this issue, we focus on robust multimodal time-series forecasting in realistic scenarios where textual information is heterogeneous and may vary in completeness, relevance, and semantic clarity. We propose PA-RNet, a perturbation-aware residual network that improves prediction reliability by extracting useful textual information while reducing the influence of noisy or misleading semantics. The key challenge is that textual information can provide valuable context for temporal variations, but may also contain incomplete, irrelevant, ambiguous, or corrupted content. Directly fusing such textual embeddings may introduce unstable semantic signals and degrade forecasting reliability.
PA-RNet is therefore designed to refine multimodal representations before fusion. The numerical branch suppresses perturbation-related components through adaptive gating, while the textual branch uses spectral residual correction to attenuate unstable semantic fluctuations and preserve stable contextual information. The refined representations are finally integrated through cross-modal attention to emphasize task-relevant textual cues aligned with numerical dynamics.

% From a theoretical perspective, we establish the Lipschitz continuity of PA-RNet with respect to textual embeddings, indicating that small textual perturbations induce only bounded variations in model outputs. We further show that the proposed spectral residual correction mechanism can reduce the expected prediction error under zero-mean textual perturbations. These theoretical results provide support for the robustness and stability of PA-RNet in noisy real-world multimodal forecasting scenarios.

% \textbf{The main contributions of this paper are as follows:}
% \begin{itemize}
%     \item We develop PA-RNet, a modular architecture that enables robust multimodal forecasting through structured denoising and modality-aware fusion. To facilitate controlled robustness evaluation, we also design a general perturbation injection pipeline that simulates realistic textual noise across benchmark datasets.
%     \item Theoretical analysis proves that the framework satisfies Lipschitz continuity with respect to textual inputs and achieves a reduction in expected prediction error under perturbations.
%     \item Extensive experiments on benchmark textual-numerical datasets demonstrate that our model achieves robust performance across varying noise levels, outperforming state-of-the-art baselines.
% \end{itemize}

\textbf{The main contributions of this paper are summarized as follows:}
\begin{itemize}
\item This paper proposes PA-RNet, a perturbation-aware residual network for robust multimodal time-series forecasting. PA-RNet integrates a numerical perturbation-aware projection module, a textual spectral residual correction module, and a cross-modal attention mechanism to learn noise-resilient multimodal representations under corrupted textual conditions.

\item This paper provides theoretical analysis showing that PA-RNet is Lipschitz continuous with respect to textual embeddings. We further demonstrate that the proposed spectral residual correction mechanism can reduce the expected prediction error under zero-mean textual perturbations.

\item Extensive experiments on benchmark textual-numerical time-series datasets show that PA-RNet consistently outperforms the strongest baselines under both original noisy textual conditions and controlled perturbation settings with varying noise intensities, demonstrating strong forecasting accuracy and robustness.
\end{itemize}

\section{Related Work}
In the domain of unimodal time series forecasting, early statistical approaches have addressed robustness against outliers and noise. For example, ~\cite{NR1} and ~\cite{NR2} proposed robust forecasting methods in 1994 and 2010, respectively, aiming to improve model stability in the presence of anomalous observations. With the rise of deep learning, time series perturbation modeling has expanded to cover adversarial attacks and noise-aware learning. Yoon et al.~\cite{NR3} introduced a robust probabilistic forecasting framework by extending traditional notions of adversarial robustness and distributional stability to the probabilistic setting, and employed randomized smoothing to construct predictors with theoretical robustness guarantees. Overall, the field of unimodal time series forecasting has established a comprehensive set of robustness modeling techniques—ranging from classical robust estimators to modern adversarial defenses.

In contrast, multimodal time series forecasting often incorporates textual modalities which, despite offering rich semantic context, are susceptible to noise, redundancy, and misleading information. While some works, like TimeXL~\cite{NR7} and NewsForecast~\cite{NR8}, utilize text refinement strategies, they mainly concentrate on filtering or selecting valuable textual content instead of explicitly enhancing the robustness of the underlying textual embedding space. As a result, there is a deficiency of model-level mechanisms designed to manage noisy or distorted textual representations. Most existing multimodal time series forecasting approaches do not incorporate denoising directly within the model architecture. The textual embeddings remain vulnerable to spurious or irrelevant signals, which may adversely affect prediction performance. 
For instance, methods like Context is Key~\cite{R10} and TimeLLM~\cite{R12} directly integrate raw text into the input using fixed prompts, without accounting for textual variability or relevance. Other approaches, such as GPT4MTS~\cite{R14} and DualTime~\cite{R16}, rely on pretrained language models to jointly model text and time series representations, yet do not explicitly filter out irrelevant or noisy textual content. As highlighted in the recent survey~\cite{R9}, TaTS~\cite{R15} and MM-TSFLib~\cite{R13} are representative intermediate fusion methods that have demonstrated strong capability in aligning modalities by leveraging pretrained models to capture cross-modal dependencies between textual descriptions and time series signals.
% These methods can be broadly categorized into early, intermediate, and late fusion strategies~\cite{R9}. Early fusion approaches, such as Context is Key~\cite{R10} and TimeLLM~\cite{R12}, integrate raw modalities at the input level using static prompts, offering simplicity but limited adaptability. Intermediate fusion methods like GPT4MTS~\cite{R14}, TaTS~\cite{R15}, and DualTime~\cite{R16} enable cross-modal interactions at the representation level, often leveraging pretrained language models for deeper integration. Late fusion methods, exemplified by MM-TSFLib~\cite{R13}, process each modality separately and merge outputs during the final prediction stage. 
While effective in aligning modalities, these methods do not explicitly address noise or redundancy in textual inputs. 
Our proposed PA-RNet performs perturbation-aware reasoning by identifying noise at the embedding level, aligning semantics across modalities through cross-modal attention. PA-RNet is theoretically shown to be Lipschitz continuous with respect to textual inputs, ensuring controlled error under noise and enhanced robustness.

\section{The PA-RNet Architecture}
This section begins by formally defining the task of multimodal time series forecasting under textual perturbations. We then present the architecture and core components of the proposed model, followed by a theoretical analysis that establishes key properties, thereby providing a solid foundation for the model’s robustness and interpretability.

\subsection{Problem Setting}
Let \( D = \{ ([x_1, s_1], \ldots, [x_n, s_n]) \} \) be a textual-numerical time series dataset, where \( x_t \in \mathbb{R} \) (for \( 1 \leq t \leq n \)) denotes the numerical value at timestamp \( t \), and \( s_t \in \mathcal{S} \) represents the textual summary associated with timestamp \( t \). The total sequence length is \( n \). 

\begin{figure*}[htbp]
    \centering
    \includegraphics[width=1\linewidth]{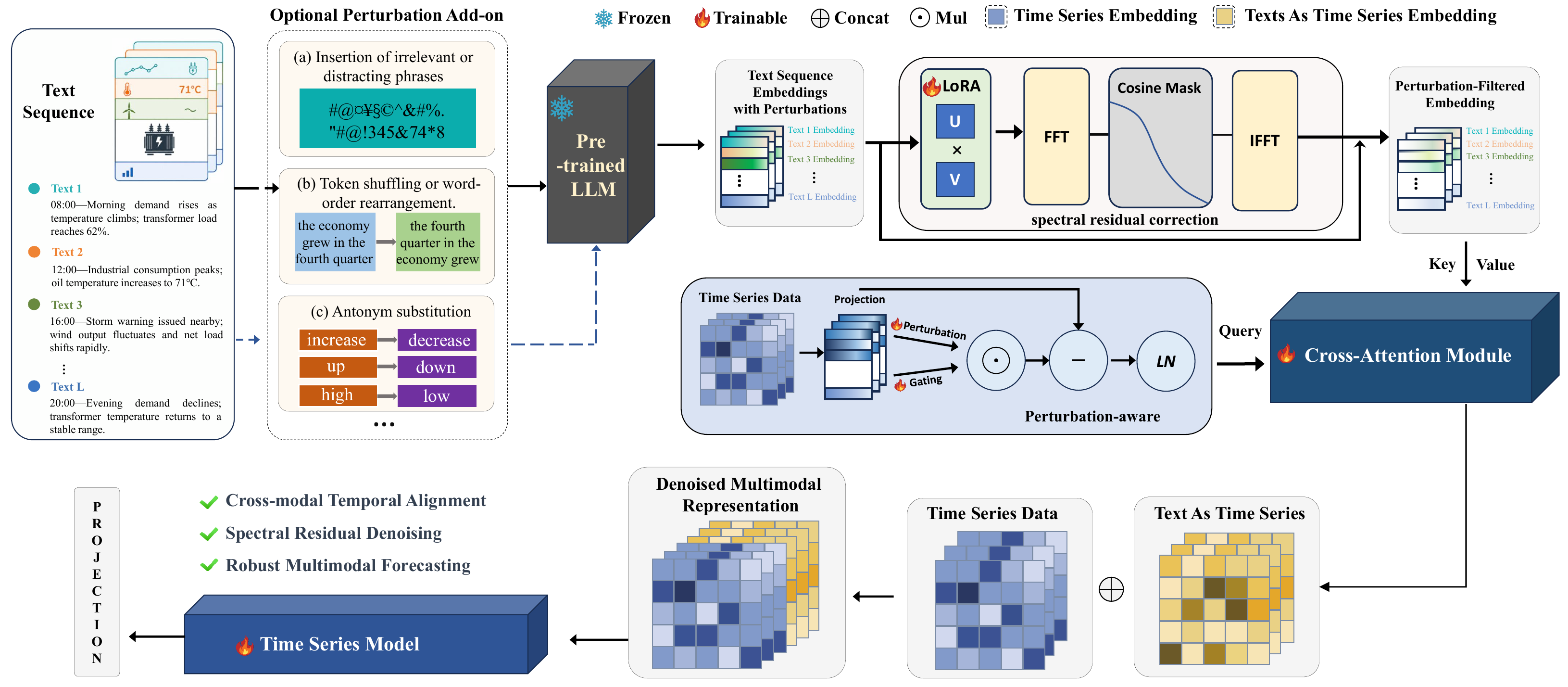}
    \caption{Overview of the PA-RNet architecture, highlighting the numerical perturbation-aware projection module, textual spectral residual correction module, and cross-modal attention mechanism.}
    \label{fig1}
\end{figure*}

Given a lookback window of length \( L \), we observe a sequence of paired inputs \( \{ ([x_1, s_1], \ldots, [x_L, s_L]) \} \), and aim to predict the next \( T \) future values \( [x_{L+1}], \ldots, [x_{L+T}] \). This task requires learning a mapping function:
\[
f: \left( [x_1, s_1], \ldots, [x_L, s_L] \right) \mapsto \left( x_{L+1}, \ldots, x_{L+T} \right),
\]
which captures the temporal dynamics of numerical values along with contextual signals from dynamic textual descriptions. To simulate real-world noise, we introduce controlled perturbations~\cite{P1}. Given a ratio $\rho \in [0,1]$, we randomly corrupt $\lfloor \rho \cdot L \rfloor$ of the past textual inputs $s_1, \ldots, s_L$ using strategies such as irrelevant phrase insertion, token shuffling, or contradiction injection. These perturbations introduce semantic distortions (see Figure~\ref{fig1}), allowing us to assess the model’s robustness under noisy textual conditions.

\subsection{Perturbation-aware Forecasting Framework}
% The overall model architecture is shown in Figure~\ref{fig2}, highlighting the interactions between the projection, attention, and forecasting components.

% The perturbation-aware projection module first maps the textual embeddings into a perturbation subspace to extract noise-sensitive components. It then computes the residual between the original embeddings and their projected counterparts, followed by a dimensionality reduction operation. This process effectively suppresses noise and redundant information, yielding cleaner semantic representations for subsequent alignment with the time-series modality. After alignment with the time-series features, the denoised textual embeddings are processed by a cross-modal attention mechanism that allows the model to selectively focus on semantically relevant textual information conditioned on temporal dynamics. 
The overall architecture of PA-RNet is illustrated in Figure~\ref{fig1}. 
PA-RNet contains a perturbation-aware numerical projection module, a spectral residual correction module, a cross-modal attention mechanism, and a forecasting backbone.

For the numerical modality, the perturbation-aware projection module learns candidate perturbation representations from the observed time-series values. 
A learnable gate adaptively controls their contribution, and the gated perturbation representation is subtracted from the original input. 
After layer normalization, the model obtains a more stable and noise-resilient numerical representation.
For the textual modality, the spectral residual correction module projects textual embeddings into a low-rank space and performs frequency-domain filtering along the temporal dimension. 
The frequency mask retains smooth semantic components while reducing unstable high-frequency fluctuations. 
The corrected textual embeddings are obtained through an inverse transformation and a residual connection.
The refined numerical representations and corrected textual embeddings are then fused by cross-modal attention, which aligns textual semantics with temporal dynamics. 
The fused multimodal representation is concatenated with the numerical input and passed to the forecasting backbone for final prediction.

\subsection{Theoretical Foundations}
For notational convenience, let 
\(x \in \mathbb{R}^{B \times L \times 1}\) denote the observed time-series values, 
where \(B\) is the batch size and \(L\) is the lookback window length. 
Let \(e_t \in \mathbb{R}^{B \times L \times D}\) denote the textual embeddings associated with the corresponding time steps, where \(D\) is the embedding dimension. 
As illustrated in Figure~\ref{fig1}, we define the proposed model as
\[
f(x,e_t)
=
F\left(
x \mathbin{\|}
A\left(
\widetilde{x},
\widetilde{e}_t
\right)
\right),
\]
where
\[
\widetilde{x}
=
\mathrm{LN}
\left(
x-\sigma(\Gamma_x(x))\odot \Phi_x(x)
\right),
\]
and
\[
\widetilde{e}_t
=
e_t
+
\mathcal{F}^{-1}
\left(
M(\omega)\odot \mathcal{F}(e_tUV)
\right).
\]
% Here, \(\Phi_x(\cdot)\) denotes the numerical projection branch, \(\Gamma_x(\cdot)\) is a learnable gating function branch, \(\sigma(\cdot)\) is the Sigmoid activation function that produces element-wise gating weights. \(U\) and \(V\) are low-rank projection matrices, \(M(\omega)\) is a frequency-domain filtering mask, and \(\mathcal{F}(\cdot)\) and \(\mathcal{F}^{-1}(\cdot)\) denote the Fourier transform and inverse Fourier transform along the temporal dimension, respectively.
Here, \(\Phi_x(\cdot)\) denotes the numerical perturbation branch, which is implemented by a fully connected network to extract perturbation-related components from the time-series modality. \(\Gamma_x(\cdot)\) represents a learnable gating branch, and the Sigmoid function \(\sigma(\cdot)\) generates element-wise gating weights to adaptively control the contribution of the projected perturbation components. For simplicity, the numerical input \(x\) is first mapped into a higher-dimensional latent space through a projection layer before being fed into these branches. \(U\) and \(V\) are low-rank projection matrices for transforming textual embeddings. \(M(\omega)\) denotes a frequency-domain filtering mask, and \(\mathcal{F}(\cdot)\) and \(\mathcal{F}^{-1}(\cdot)\) denote the Fourier transform and inverse Fourier transform along the temporal dimension, respectively.

For the \(k\)-th frequency component, \(M(\omega)\) is defined as
\[
M_k =
\begin{cases}
1, & 0 \leq k < k_c, \\[4pt]
\frac{1}{2}
\left(
1+\cos\left(
\pi \frac{k-k_c}{k_e-k_c}
\right)
\right), 
& k_c \leq k < k_e, \\[6pt]
0, & k_e \leq k < K,
\end{cases}
\]
% where \(K\) is the number of frequency components, \(k_c\) is the cutoff frequency index, and \(k_e\) denotes the end index of the transition region. 
To determine the cutoff boundary and the transition region in the frequency domain, we compute the cutoff frequency index and the transition end index as follows:
\[
k_c = \left\lfloor K \cdot r_c \right\rfloor,
\]
\[
k_e = \min \left( k_c + \max \left(1, \left\lfloor K \cdot r_t \right\rfloor \right), K \right),
\]
where \(K\) denotes the number of frequency components, \(r_c\) is the cutoff frequency ratio, and \(r_t\) is the transition-width ratio. The resulting \(k_c\) is a discrete cutoff frequency index, while \(k_e\) denotes the end index of the transition region. The floor operation converts the ratio-based cutoff position into an integer frequency index.

Based on the above formulation, we provide two theoretical results to characterize the robustness of PA-RNet under noisy textual inputs. Specifically, Proposition~\ref{prop:lipschitz} establishes the Lipschitz continuity of PA-RNet with respect to textual embeddings, while Proposition~\ref{prop:error_reduction} shows that the proposed spectral residual correction can reduce the expected prediction error under zero-mean textual perturbations. The detailed proofs are provided in Appendix~\ref{app:theoretical_proofs}.

\paragraph{Simulated Textual Noise.}
To evaluate the robustness of PA-RNet under perturbed textual inputs, we introduce several simple and controllable noise operations, including irrelevant phrase insertion, token shuffling, antonym substitution, text truncation, and number generalization. These perturbations are not intended to exhaustively model all real-world textual corruptions, but rather to provide controlled stress tests that reflect typical forms of noisy, incomplete, or semantically shifted textual information. We then apply different perturbation types and noise intensities to the original textual inputs to assess whether the model maintains stable forecasting performance under noisy-text conditions.

\begin{table*}[ht]
\centering
\caption{Statistics of Datasets from Nine Real-world Domains in Time-MMD Benchmark}
% \footnotesize
\resizebox{\textwidth}{!}{
\begin{tabular}{lccccccc}
\toprule
\textbf{Domain} & \textbf{Target} & \textbf{Dimension} & \textbf{Frequency} & \textbf{Samples} & \textbf{Timespan} & \textbf{Lookback} & \textbf{Horizon} \\
\midrule
Agriculture   & Broiler Composite                  & 1  & Monthly & 496  & 1983 -- Present & 8 &\{6, 8, 10, 12\} \\
Climate       & Drought Level                      & 5  & Monthly & 496  & 1983 -- Present & 8 & \{6, 8, 10, 12\} \\
Economy       & International Trade Balance        & 3  & Monthly & 423  & 1989 -- Present & 8 & \{6, 8, 10, 12\} \\
Energy        & Gasoline Prices                    & 9  & Weekly  & 1479 & 1996 -- Present & 36 & \{12, 24, 36, 48\} \\
Environment   & Air Quality Index                  & 4  & Daily   & 11102& 1982 -- 2023    & 96 & \{48, 96, 192, 336\} \\
Health        & Influenza Patients Proportion      & 11 & Weekly  & 1389 & 1997 -- Present & 36 & \{12, 24, 36, 48\} \\
Security      & Disaster and Emergency Grants      & 1  & Monthly & 297  & 1999 -- Present & 8 & \{6, 8, 10, 12\} \\
Social Good   & Unemployment Rate                  & 1  & Monthly & 900  & 1950 -- Present & 8 & \{6, 8, 10, 12\} \\
Traffic       & Travel Volume                      & 1  & Monthly & 531  & 1980 -- Present & 8 & \{6, 8, 10, 12\} \\
\bottomrule
\end{tabular}
}
\label{tab:dataset_stats}
\end{table*}

\section{Experiments} 
\subsection{Datasets}
To evaluate the generalizability and robustness of our model, we conduct experiments on the benchmark Time-MMD dataset suite~\cite{R13}, which spans 9 real-world domains with diverse temporal resolutions, including monthly, weekly, and daily observations. Notably, the textual modality in Time-MMD is derived from real-world textual sources and is not perfectly aligned with the numerical time series in all cases. It may contain incomplete descriptions, partially irrelevant information, or noisy contextual signals. Therefore, the dataset provides a realistic testbed for assessing whether multimodal forecasting models can effectively exploit textual information under imperfect textual conditions. Further details of the datasets, including dataset 
% statistics, are provided in Appendix~\ref{tab:dataset_stats}.
statistics, are provided in Table~\ref{tab:dataset_stats}.

To avoid information leakage, we follow the preprocessing strategy of TaTS~\cite{R15}, ensuring that textual events used as inputs occur no later than their corresponding time steps.

\subsection{Baseline Methods}
To handle the textual modality, we employ a GPT2-based~\cite{gpt2} encoder to extract textual representations from text inputs under different types and levels of perturbations. After obtaining the textual embeddings, the proposed PA-RNet framework can be flexibly combined with existing time-series forecasting models. To evaluate its compatibility and generality, we integrate PA-RNet with nine widely used backbones from different categories, including:
(i) Transformer-based models: iTransformer~\cite{R6}, PatchTST~\cite{R5}, Crossformer~\cite{base1}, Autoformer~\cite{R17}, Informer~\cite{R7}, and Transformer~\cite{c:22}; 
(ii) Linear models: DLinear~\cite{R4}; 
and (iii) Frequency-based models: FEDformer~\cite{R8} and FiLM~\cite{base2}. 

For each backbone model, we compare PA-RNet with three representative baselines. 

(i) \textbf{Numerical-only}: a uni-modal forecasting setting that ignores the paired textual information and only uses numerical time-series inputs. 

(ii) \textbf{MM-TSFLib}~\cite{R13}: a recently proposed multimodal time-series forecasting library that incorporates textual information by linearly interpolating the outputs of the time-series model with text embeddings, while treating texts in a bag-of-words manner. 

(iii) \textbf{TaTS}~\cite{R15}: an architecture for multimodal time-series forecasting that embeds external texts as auxiliary temporal variables to assist prediction.
% (iv)Foundation-model-based time-series forecasting methods:TimeCMA and Time-VLM

Beyond these backbone-compatible baselines, we further include representative (iv) \textbf{Foundation-model-based methods}: representative methods including \textbf{TimeCMA}~\cite{base3} and \textbf{Time-VLM}~\cite{base4}, which leverage pre-trained LLMs or VLMs for multimodal time-series forecasting.

\subsection{Experimental Setup}
The experiments were conducted on a Linux system with an x86\_64 CPU and 251.38 GB RAM, utilizing NVIDIA Tesla V100 GPUs (32 GB) for computation. The software environment is based on Python 3.11.11 managed via conda, with key packages including \texttt{numpy}, \texttt{pandas}, \texttt{torch}, \texttt{transformers}, and others.

\subsection{Evaluation Metrics}

To comprehensively evaluate the performance of our model on the multimodal time series forecasting task, we adopt two widely used metrics: Mean Squared Error (MSE)~\cite{MC1} and Mean Absolute Error (MAE)~\cite{MC2}. To ensure the reliability of the experimental results, all experiments are repeated with five different random seeds.

\begin{table*}[t]
\centering
\scriptsize
\setlength{\tabcolsep}{2.0pt}
\renewcommand{\arraystretch}{1.05}
\caption{Forecasting performance on benchmark textual-numerical time-series datasets. Lower MSE/MAE indicates better performance. Bold values indicate the best results or results tied with the best-performing baseline.}
\label{tab:benchmark_results}
\resizebox{\textwidth}{!}{
\begin{tabular}{ll|cc|cc|cc|cc|cc|cc|cc|cc|cc}
\toprule
\multirow{2}{*}{Dataset} & \multirow{2}{*}{Method} & \multicolumn{2}{c|}{Autoformer} & \multicolumn{2}{c|}{Crossformer} & \multicolumn{2}{c|}{DLinear} & \multicolumn{2}{c|}{FEDformer} & \multicolumn{2}{c|}{FiLM} & \multicolumn{2}{c|}{Informer} & \multicolumn{2}{c|}{iTransformer} & \multicolumn{2}{c|}{PatchTST} & \multicolumn{2}{c}{Transformer} \\
\cmidrule(lr){3-4} \cmidrule(lr){5-6} \cmidrule(lr){7-8} \cmidrule(lr){9-10} \cmidrule(lr){11-12} \cmidrule(lr){13-14} \cmidrule(lr){15-16} \cmidrule(lr){17-18} \cmidrule(lr){19-20}
 & & MSE & MAE & MSE & MAE & MSE & MAE & MSE & MAE & MSE & MAE & MSE & MAE & MSE & MAE & MSE & MAE & MSE & MAE \\
\midrule
\multirow{5}{*}{Agriculture} & Uni-modal & 0.103 & 0.235 & 0.309 & 0.405 & 0.195 & 0.334 & 0.100 & 0.234 & 0.100 & 0.198 & 0.414 & 0.471 & 0.096 & 0.209 & 0.093 & 0.197 & 0.283 & 0.367 \\
 & MM-TSFLib & 0.103 & 0.236 & 0.248 & 0.350 & \textbf{0.144} & \textbf{0.259} & 0.101 & 0.232 & 0.100 & 0.198 & 0.342 & 0.419 & 0.094 & 0.205 & 0.092 & 0.196 & 0.230 & 0.328 \\
 & TaTS & 0.104 & 0.237 & 0.177 & 0.287 & 0.152 & 0.277 & 0.100 & 0.236 & 0.099 & 0.196 & 0.214 & 0.319 & 0.094 & 0.204 & 0.092 & 0.195 & 0.158 & 0.282 \\
 & \textbf{PA-RNet} & \textbf{0.102} & \textbf{0.233} & \textbf{0.172} & \textbf{0.276} & 0.150 & 0.276 & \textbf{}\textbf{0.100} & \textbf{0.228} & \textbf{0.099} & \textbf{0.196} & \textbf{0.206} & \textbf{0.313} & \textbf{0.094} & \textbf{0.204} & \textbf{0.091} & \textbf{0.195} & \textbf{0.153} & \textbf{0.280} \\ 
\midrule
\multirow{5}{*}{Climate} & Uni-modal & 1.289 & 0.919 & 1.165 & 0.854 & 1.283 & 0.921 & 1.151 & 0.864 & 1.368 & 0.961 & 1.168 & 0.850 & 1.199 & 0.900 & 1.296 & 0.933 & 1.062 & 0.807 \\
 & MM-TSFLib & 1.133 & 0.862 & 1.123 & 0.835 & 1.103 & 0.850 & 1.037 & 0.820 & 1.191 & 0.894 & 1.107 & 0.827 & 1.106 & 0.862 & 1.147 & 0.873 & 1.035 & 0.795 \\
 & TaTS & 0.993 & 0.804 & 1.033 & 0.789 & 0.950 & 0.778 & 0.964 & 0.788 & 0.986 & 0.797 & 1.017 & 0.788 & 1.018 & 0.809 & 1.001 & 0.800 & \textbf{0.956} & \textbf{0.761} \\
 & \textbf{PA-RNet} & \textbf{0.972} & \textbf{0.797} & \textbf{1.015} & \textbf{0.782} & \textbf{0.948} & \textbf{0.777} & \textbf{0.955} & \textbf{0.787} & \textbf{0.983} & \textbf{0.795} & \textbf{0.970} & \textbf{0.773} & \textbf{1.016} & \textbf{0.807} & \textbf{0.992} & \textbf{0.796} & 0.972 & 0.764 \\
\midrule
\multirow{5}{*}{Economy} & Uni-modal & 0.083 & 0.227 & 0.566 & 0.642 & 0.113 & 0.281 & 0.077 & 0.216 & 0.018 & 0.108 & 0.572 & 0.712 & 0.016 & 0.098 & 0.019 & 0.107 & 0.129 & 0.276 \\
 & MM-TSFLib & 0.083 & 0.223 & 0.380 & 0.517 & 0.050 & 0.176 & 0.070 & 0.207 & 0.018 & 0.108 & 0.490 & 0.654 & 0.016 & 0.098 & 0.019 & 0.107 & 0.167 & 0.335 \\
 & TaTS & 0.028 & 0.132 & 0.172 & 0.363 & 0.016 & 0.102 & 0.036 & 0.145 & 0.010 & 0.085 & 0.310 & 0.514 & 0.011 & 0.085 & 0.011 & 0.086 & 0.133 & 0.311 \\
 & \textbf{PA-RNet} & \textbf{0.024} & \textbf{0.121} & \textbf{0.155} & \textbf{0.338} & \textbf{0.016} & \textbf{0.102} & \textbf{0.035} & \textbf{0.142} & \textbf{0.010} & \textbf{0.085} & \textbf{0.300} & \textbf{0.494} & \textbf{0.010} & \textbf{0.084} & \textbf{0.011} & \textbf{0.086} & \textbf{0.104} & \textbf{0.267} \\
\midrule
\multirow{5}{*}{Energy} & Uni-modal & 0.357 & 0.461 & 0.331 & 0.429 & 0.257 & 0.358 & 0.224 & 0.342 & 0.261 & 0.367 & 0.387 & 0.476 & 0.288 & 0.403 & 0.300 & 0.414 & 0.369 & 0.456 \\
 & MM-TSFLib & 0.351 & 0.459 & 0.320 & 0.419 & 0.254 & 0.359 & \textbf{0.222} & \textbf{0.339} & \textbf{0.257} & \textbf{0.364} & 0.383 & 0.479 & 0.282 & 0.400 & 0.294 & 0.406 & 0.361 & 0.461 \\
 & TaTS & \textbf{0.344} & \textbf{0.452} & 0.335 & 0.428 & \textbf{0.252} & 0.358 & 0.239 & 0.364 & 0.261 & 0.368 & 0.377 & 0.461 & 0.296 & 0.408 & 0.292 & 0.408 & 0.321 & \textbf{0.422} \\
 & \textbf{PA-RNet} & 0.347 & 0.455 & \textbf{0.316} & \textbf{0.417} & 0.255 & \textbf{0.357} & 0.240 & 0.363 & 0.259 & 0.367 & \textbf{0.355} & \textbf{0.461} &\textbf{0.279} & \textbf{0.398} & \textbf{0.288} & \textbf{0.400} & \textbf{0.315} & 0.424 \\
\midrule
\multirow{5}{*}{Environment} & Uni-modal & 0.441 & 0.528 & 0.444 & 0.524 & 0.532 & 0.619 & 0.386 & 0.484 & 0.320 & 0.392 & 0.438 & 0.522 & 0.278 & 0.389 & 0.330 & 0.411 & 0.384 & 0.481 \\
 & MM-TSFLib & 0.371 & 0.473 & 0.354 & 0.447 & 0.323 & 0.423 & 0.272 & 0.387 & 0.274 & 0.373 & 0.388 & 0.481 & 0.284 & 0.399 & 0.301 & 0.406 & 0.345 & 0.447 \\
 & TaTS & 0.312 & 0.421 & 0.298 & 0.418 & 0.308 & 0.441 & \textbf{0.282} & \textbf{0.389} & \textbf{0.268} & \textbf{0.369} & 0.291 & 0.409 & 0.260 & \textbf{0.368} & 0.269 & 0.373 & 0.280 & 0.394 \\
 & \textbf{PA-RNet} & \textbf{0.308} & \textbf{0.417} & \textbf{0.298} & \textbf{0.416} & \textbf{0.308} & \textbf{0.440} & 0.284 & 0.393 & 0.269 & 0.370 & \textbf{0.287} & \textbf{0.400} & \textbf{0.259} & 0.371 & \textbf{0.269} & \textbf{0.373} & \textbf{0.276} & \textbf{0.392} \\
\midrule
\multirow{5}{*}{Health} & Uni-modal & 1.810 & 0.993 & 1.423 & 0.790 & 1.664 & 0.885 & 1.487 & 0.928 & 1.563 & 0.851 & 1.522 & 0.818 & 1.825 & 0.806 & 1.645 & 0.788 & 1.460 & 0.798 \\
 & MM-TSFLib & 1.717 & 0.948 & 1.389 & 0.775 & 1.568 & 0.854 & 1.391 & 0.876 & 1.463 & 0.818 & 1.380 & 0.776 & 1.613 & 0.781 & 1.442 & 0.753 & 1.287 & 0.755 \\
 & TaTS & 1.491 & 0.883 & 1.239 & 0.737 & 1.433 & \textbf{0.810} & 1.352 & 0.847 & 1.350 & 0.784 & 1.178 & 0.731 & 1.375 & 0.748 & 1.260 & 0.725 & \textbf{1.131} & 0.705 \\
 & \textbf{PA-RNet} & \textbf{1.466 }& \textbf{0.869} & \textbf{1.237} & \textbf{0.734} & \textbf{1.429} & 0.812 & \textbf{1.338} & \textbf{0.840} & \textbf{1.345} & \textbf{0.782} & \textbf{1.163} & \textbf{0.729} & \textbf{1.339} & \textbf{0.748} & \textbf{1.252} & \textbf{0.724} & 1.139 & \textbf{0.702} \\
\midrule
\multirow{5}{*}{Security} & Uni-modal & 125.196 & 5.419 & 126.382 & 6.244 & 109.359 & 5.196 & 133.267 & 5.668 & 129.269 & 6.259 & 130.349 & 6.578 & 108.577 & 4.743 & 125.985 & 6.014 & 126.582 & 6.271 \\
 & MM-TSFLib & 125.277 & 5.510 & 128.397 & 6.410 & 107.439 & 4.894 & 131.983 & 5.642 & 119.393 & 5.731 & 130.325 & 6.579 & 107.900 & 4.837 & 126.730 & 6.046 & 127.868 & 6.329 \\
 & TaTS & 110.493 & 4.937 & 126.387 & 6.244 & 107.388 & 4.892 & 108.862 & 4.917 & 113.517 & \textbf{5.166} & \textbf{124.675} & \textbf{6.185} & \textbf{106.983} & \textbf{4.803} & 125.607 & 6.006 & 122.698 & 5.857 \\
 & \textbf{PA-RNet} & \textbf{110.141} & \textbf{4.918} & \textbf{126.080} & \textbf{6.220} & \textbf{107.383} & \textbf{4.892} & \textbf{108.691} & \textbf{4.914} & \textbf{113.486} & 5.167 & 124.998 & 6.209 & 107.554 & 4.816 & \textbf{124.277} & \textbf{5.980} & \textbf{121.850} & \textbf{5.807} \\
\midrule
\multirow{5}{*}{Social Good} & Uni-modal & 1.128 & \textbf{0.593} & 0.854 & 0.471 & 1.041 & 0.574 & 1.180 & \textbf{0.527} & 1.118 & 0.524 & 0.805 & 0.422 & 1.164 & 0.593 & 1.052 & 0.517 & 0.819 & 0.427 \\
 & MM-TSFLib & 1.087 & 0.600 & 0.846 & 0.477 & 1.025 & 0.579 & 1.128 & 0.541 & 1.040 & 0.537 & \textbf{0.800} & 0.432 & \textbf{1.047} & \textbf{0.535} & 1.037 & 0.544 & 0.825 & 0.435 \\
 & TaTS & \textbf{1.066} & 0.598 & 0.849 & 0.471 & 1.034 & 0.579 & 1.081 & 0.575 & 1.037 & 0.520 & 0.827 & 0.423 & 1.129 & 0.680 & 1.036 & \textbf{0.512} & 0.832 & 0.427 \\
 & \textbf{PA-RNet} & 1.068 & 0.597 & \textbf{0.846} & \textbf{0.456} & \textbf{1.016} & \textbf{0.569} & \textbf{1.079} & 0.571 & \textbf{1.033} & \textbf{0.515} & 0.810 & \textbf{0.411} & 1.136 & 0.634 & \textbf{1.028} & 0.522 & \textbf{0.818} & \textbf{0.420} \\
\midrule
\multirow{5}{*}{Traffic} & Uni-modal & 0.194 & 0.273 & 0.201 & 0.318 & 0.325 & 0.438 & 0.203 & 0.274 & 0.254 & 0.330 & 0.163 & 0.295 & 0.194 & 0.242 & 0.194 & 0.242 & 0.163 & 0.273 \\
 & MM-TSFLib & 0.181 & 0.263 & 0.194 & 0.312 & 0.243 & 0.364 & 0.187 & 0.256 & 0.237 & 0.317 & 0.171 & 0.303 & 0.194 & 0.250 & 0.187 & 0.244 & 0.165 & 0.274 \\
 & TaTS & 0.162 & 0.224 & 0.170 & 0.285 & 0.197 & 0.311 & 0.171 & 0.228 & 0.183 & 0.258 & \textbf{0.152} & 0.271 & 0.182 & 0.222 & 0.176 & 0.219 & \textbf{0.151} & 0.256 \\
 & \textbf{PA-RNet} & \textbf{0.161} & \textbf{0.224} & \textbf{0.169} & \textbf{0.283} & \textbf{0.196} & \textbf{0.310} & \textbf{0.169} & \textbf{0.227} & \textbf{0.183} & \textbf{0.257} & 0.156 & \textbf{0.265} & \textbf{0.178} & \textbf{0.221} & \textbf{0.175} & \textbf{0.219} & 0.155 & \textbf{0.244} \\
\midrule
\multicolumn{2}{c|}{\textbf{Avg. Rank} $\downarrow$} &
\multicolumn{18}{c}{
\textbf{PA-RNet}: \textbf{1.28} \quad
TaTS: 1.99 \quad
MM-TSFLib: 2.99 \quad
Uni-modal: 3.51
} \\
\bottomrule
\end{tabular}
}
\end{table*}

\begin{table*}[t]
\centering
\scriptsize
\setlength{\tabcolsep}{2.0pt}
\renewcommand{\arraystretch}{1.05}
\caption{Average robustness results under textual perturbations on benchmark textual-numerical time-series datasets. For MM-TSFLib, TaTS, and PA-RNet, results are averaged over perturbation ratios $\rho\in\{0.3,0.5,0.7,0.9\}$. Bold values indicate the best results or results tied with the best-performing baseline.}
\label{tab:perturbation_avg_results}
\resizebox{\textwidth}{!}{
\begin{tabular}{ll|cc|cc|cc|cc|cc|cc|cc|cc|cc}
\toprule
\multirow{2}{*}{Dataset} & \multirow{2}{*}{Method} & \multicolumn{2}{c|}{Autoformer} & \multicolumn{2}{c|}{Crossformer} & \multicolumn{2}{c|}{DLinear} & \multicolumn{2}{c|}{FEDformer} & \multicolumn{2}{c|}{FiLM} & \multicolumn{2}{c|}{Informer} & \multicolumn{2}{c|}{iTransformer} & \multicolumn{2}{c|}{PatchTST} & \multicolumn{2}{c}{Transformer} \\
\cmidrule(lr){3-4} \cmidrule(lr){5-6} \cmidrule(lr){7-8} \cmidrule(lr){9-10} \cmidrule(lr){11-12} \cmidrule(lr){13-14} \cmidrule(lr){15-16} \cmidrule(lr){17-18} \cmidrule(lr){19-20}
 & & MSE & MAE & MSE & MAE & MSE & MAE & MSE & MAE & MSE & MAE & MSE & MAE & MSE & MAE & MSE & MAE & MSE & MAE \\
\midrule
\multirow{5}{*}{Agriculture} & Uni-modal & \textbf{0.103} & \textbf{0.235} & 0.309 & 0.405 & 0.195 & 0.334 & 0.100 & 0.234 & 0.100 & 0.198 & 0.414 & 0.471 & 0.096 & 0.209 & 0.093 & 0.197 & 0.283 & 0.367 \\
 & MM-TSFLib & 0.103 & 0.236 & 0.247 & 0.349 & \textbf{0.144} & \textbf{0.260} & 0.101 & 0.232 & 0.100 & 0.198 & 0.345 & 0.422 & 0.103 & 0.232 & 0.092 & 0.196 & 0.232 & 0.331 \\
 & TaTS & 0.104 & 0.235 & 0.177 & 0.287 & 0.152 & 0.277 & 0.101 & 0.237 & 0.099 & 0.196 & \textbf{0.214} & \textbf{0.319} & 0.094 & \textbf{0.205} & 0.092 & 0.195 & 0.158 & 0.282 \\
 & \textbf{PA-RNet} & 0.104 & 0.236 & \textbf{0.172} & \textbf{0.276} & 0.150 & 0.276 & \textbf{0.100} & \textbf{0.229} & \textbf{0.099} & \textbf{0.196} & 0.218 & 0.320 & \textbf{0.094} & 0.207 & \textbf{0.091} & \textbf{0.195} & \textbf{0.158} & \textbf{0.281} \\
\midrule
\multirow{5}{*}{Climate} & Uni-modal & 1.289 & 0.919 & 1.165 & 0.854 & 1.283 & 0.921 & 1.151 & 0.864 & 1.368 & 0.961 & 1.168 & 0.850 & 1.199 & 0.900 & 1.296 & 0.933 & 1.062 & 0.807 \\
 & MM-TSFLib & 1.135 & 0.861 & 1.124 & 0.835 & 1.102 & 0.850 & 1.041 & 0.820 & 1.192 & 0.895 & 1.113 & 0.829 & 1.105 & 0.861 & 1.151 & 0.874 & 1.036 & 0.796 \\
 & TaTS & 1.001 & 0.808 & 1.033 & 0.789 & 0.950 & 0.778 & 0.961  & 0.788 & 0.986 & 0.797 & 1.012 & 0.786 & 1.020 & 0.809 & 1.001 & 0.800 & \textbf{0.956} & \textbf{0.761} \\
 & \textbf{PA-RNet} & \textbf{0.972} & \textbf{0.797} & \textbf{1.015} & \textbf{0.782} & \textbf{0.948} & \textbf{0.777} & \textbf{0.952} & \textbf{0.788} & \textbf{0.983} & \textbf{0.795} & \textbf{0.975} & \textbf{0.775} & \textbf{1.014} & \textbf{0.806} & \textbf{0.992} & \textbf{0.796} & 0.974 & 0.765 \\
\midrule
\multirow{5}{*}{Economy} & Uni-modal & 0.083 & 0.227 & 0.566 & 0.642 & 0.113 & 0.281 & 0.077 & 0.216 & 0.018 & 0.108 & 0.572 & 0.712 & 0.016 & 0.098 & 0.019 & 0.107 & 0.129 & 0.276 \\
 & MM-TSFLib & 0.085 & 0.225 & 0.405 & 0.546 & 0.051 & 0.177 & 0.070 & 0.207 & 0.018 & 0.108 & 0.482 & 0.648 & 0.016 & 0.098 & 0.019 & 0.107 & 0.166 & 0.334 \\
 & TaTS & 0.026 & 0.128 & 0.162 & 0.347 & 0.016 & 0.102 & 0.036 & 0.145 & 0.010 & 0.085 & 0.307 & 0.511 & 0.011 & 0.085 & 0.011 & 0.086 & 0.133 & 0.311 \\
 & \textbf{PA-RNet} & \textbf{0.023} & \textbf{0.121} & \textbf{0.158} & \textbf{0.346} & \textbf{0.016} & \textbf{0.102} & \textbf{0.035} & \textbf{0.142} & \textbf{0.010} & \textbf{0.085} & \textbf{0.300} & \textbf{0.496} & \textbf{0.010} & \textbf{0.084} & \textbf{0.011} & \textbf{0.086} & \textbf{0.120} & \textbf{0.286} \\
\midrule
\multirow{5}{*}{Health} & Uni-modal & 1.810 & 0.993 & 1.423 & 0.790 & 1.664 & 0.885 & 1.487 & 0.928 & 1.563 & 0.851 & 1.522 & 0.818 & 1.825 & 0.806 & 1.645 & 0.788 & 1.460 & 0.798 \\
 & MM-TSFLib & 1.723 & 0.954 & 1.433 & 0.789 & 1.568 & 0.854 & 1.398 & 0.881 & 1.463 & 0.819 & 1.388 & 0.777 & 1.615 & 0.782 & 1.444 & 0.754 & 1.308 & 0.757 \\
 & TaTS & 1.496 & 0.883 & 1.257 & 0.743 & 1.433 & \textbf{0.810} & 1.347 & 0.846 & 1.350 & 0.784 & 1.170 & 0.727 & 1.380 & \textbf{0.749} & 1.260 & 0.725 & \textbf{1.135} & 0.706 \\
 & \textbf{PA-RNet} & \textbf{1.475} & \textbf{0.874} & \textbf{1.243} & \textbf{0.738} & \textbf{1.429} & 0.812 & \textbf{1.342} & \textbf{0.845} & \textbf{1.345} & \textbf{0.782} & \textbf{1.155} & \textbf{0.727} & \textbf{1.369} & 0.751 & \textbf{1.252} & \textbf{0.724} & 1.146 & \textbf{0.704} \\
\midrule
\multirow{5}{*}{Security} & Uni-modal & 125.196 & 5.419 & 126.382 & 6.244 & 109.359 & 5.196 & 133.267 & 5.668 & 129.269 & 6.259 & 130.349 & 6.578 & 108.059 & 4.951 & 125.985 & 6.014 & 126.582 & 6.271 \\
 & MM-TSFLib & 125.794 & 5.522 & 128.400 & 6.410 & 107.438 & 4.893 & 131.937 & 5.639 & 119.407 & 5.732 & 128.135 & 6.420 & 107.685 & 4.819 & 127.010 & 6.054 & 128.811 & 6.391 \\
 & TaTS & 110.176 & 4.927 & 126.387 & 6.244 & 107.388 & 4.892 & 108.861 & 4.917 & 113.517 & \textbf{5.166} & \textbf{124.678} & \textbf{6.185} & \textbf{107.196} & \textbf{4.825} & 125.607 & 6.006 & \textbf{122.107} & \textbf{5.802} \\
 & \textbf{PA-RNet} & \textbf{110.103} & \textbf{4.919} & \textbf{126.084} & \textbf{6.220} & \textbf{107.386} & \textbf{4.892} & \textbf{108.691} & \textbf{4.914} & \textbf{113.486} & 5.167 & 124.725 & 6.186 & 107.486 & 4.886 & \textbf{124.277} & \textbf{5.980} & 122.885 & 5.887 \\
\midrule
\multirow{5}{*}{Traffic} & Uni-modal & 0.194 & 0.273 & 0.201 & 0.318 & 0.325 & 0.438 & 0.203 & 0.274 & 0.254 & 0.330 & 0.163 & 0.295 & 0.194 & 0.242 & 0.193 & 0.242 & 0.163 & 0.273 \\
 & MM-TSFLib & 0.185 & 0.267 & 0.194 & 0.312 & 0.245 & 0.367 & 0.188 & 0.259 & 0.237 & 0.316 & 0.170 & 0.302 & 0.187 & 0.242 & 0.187 & 0.245 & 0.165 & 0.273 \\
 & TaTS & 0.164 & 0.225 & 0.170 & 0.285 & 0.197 & 0.311 & 0.171 & \textbf{0.228} & 0.183 & 0.258 & \textbf{0.152} & 0.270 & 0.183 & 0.223 & 0.176 & 0.219 & \textbf{0.151} & 0.256 \\
 & \textbf{PA-RNet} & \textbf{0.161} & \textbf{0.224} & \textbf{0.169} & \textbf{0.283} & \textbf{0.196} & \textbf{0.310} & \textbf{0.170} & 0.229 & \textbf{0.183} & \textbf{0.257} & 0.153 & \textbf{0.261} & \textbf{0.178} & \textbf{0.221} & \textbf{0.175} & \textbf{0.219} & 0.154 & \textbf{0.248} \\
\midrule
\multicolumn{2}{c|}{\textbf{Avg. Rank} $\downarrow$} &
\multicolumn{18}{c}{
\textbf{PA-RNet}: \textbf{1.24} \quad
TaTS: 1.77 \quad
MM-TSFLib: 3.16 \quad
Uni-modal: 3.68
} \\
\bottomrule
\end{tabular}
}
\end{table*}

\begin{table*}[htbp]
\centering
\footnotesize
\caption{Ablation study of PA-RNet on Economy, Health, and Agriculture datasets. The best results are highlighted in \textbf{bold}, and the second-best results are \underline{underlined}.}
\label{tab:ablation}
\resizebox{\textwidth}{!}{
\begin{tabular}{llcccccccc}
\toprule
\multirow{2}{*}{Dataset} & \multirow{2}{*}{Horizon}
& \multicolumn{2}{c}{w/o Cross-Attention}
& \multicolumn{2}{c}{w/o spectral residual}
& \multicolumn{2}{c}{w/o perturbation-aware}
& \multicolumn{2}{c}{PA-RNet} \\
\cmidrule(lr){3-4} \cmidrule(lr){5-6} \cmidrule(lr){7-8} \cmidrule(lr){9-10}
& & MSE & MAE & MSE & MAE & MSE & MAE & MSE & MAE \\
\midrule

\multirow{5}{*}{Economy}
& 6  & 0.114 & 0.295 & 0.112 & 0.286 & \underline{0.087} & \underline{0.248} & \textbf{0.072} & \textbf{0.226} \\
& 8  & 0.131 & 0.309 & 0.138 & 0.319 & \underline{0.129} & \textbf{0.304} & \textbf{0.120} & \underline{0.305} \\
& 10 & 0.223 & 0.425 & 0.238 & 0.442 & \underline{0.220} & \underline{0.420} & \textbf{0.188} & \textbf{0.383} \\
& 12 & \textbf{0.215} & \textbf{0.405} & 0.259 & 0.466 & \underline{0.230} & \underline{0.433} & 0.240 & 0.439 \\
& Avg & 0.171 & 0.359 & 0.187 & 0.378 & \underline{0.167} & \underline{0.351} & \textbf{0.155} & \textbf{0.338} \\
\midrule

\multirow{5}{*}{Health}
& 12 & \underline{1.026} & \underline{0.672} & 1.036 & 0.692 & 1.135 & 0.713 & \textbf{0.971} & \textbf{0.661} \\
& 24 & 1.464 & 0.802 & 1.391 & 0.783 & \textbf{1.268} & \textbf{0.742} & \underline{1.297} & \underline{0.751} \\
& 36 & 1.280 & 0.744 & \underline{1.223} & \underline{0.722} & \textbf{1.206} & \textbf{0.719} & 1.294 & 0.747 \\
& 48 & \underline{1.378} & 0.785 & \textbf{1.324} & \textbf{0.762} & 1.495 & 0.815 & 1.384 & \underline{0.779} \\
& Avg & 1.287 & 0.751 & \underline{1.244} & \underline{0.740} & 1.276 & 0.747 & \textbf{1.237} & \textbf{0.734} \\
\midrule

\multirow{5}{*}{Agriculture}
& 6  & 0.104 & \textbf{0.220} & \underline{0.098} & \textbf{0.220} & 0.100 & 0.231 & \textbf{0.093} & \textbf{0.220} \\
& 8  & 0.161 & 0.282 & 0.155 & 0.270 & \underline{0.148} & \textbf{0.258} & \textbf{0.146} & \underline{0.261} \\
& 10 & \underline{0.195} & 0.298 & 0.201 & 0.311 & \textbf{0.184} & \textbf{0.286} & 0.197 & \underline{0.293} \\
& 12 & \textbf{0.245} & \underline{0.339} & 0.254 & 0.351 & 0.266 & 0.350 & \underline{0.252} & \textbf{0.329} \\
& Avg & 0.176 & 0.285 & 0.177 & 0.288 & \underline{0.175} & \underline{0.281} & \textbf{0.172} & \textbf{0.276} \\
\bottomrule
\end{tabular}
}
\end{table*}

\subsection{Experimental Results and Insights}

The experimental results under the original noisy textual conditions are reported in Table~\ref{tab:benchmark_results}. Overall, PA-RNet achieves the best or tied-best performance in 129 out of 162 evaluation entries, demonstrating its strong effectiveness on benchmark textual-numerical time-series datasets. In the table, bold font indicates results that are either the best or tied with the strongest competing baseline. We also report the average rank of each method across all datasets, backbones, and metrics, where a lower rank indicates better overall performance~\cite{rank1,rank2}.

We further evaluate the robustness of PA-RNet under different levels of injected textual perturbations. Specifically, the perturbation ratio \(\rho\) is set to 0.3, 0.5, 0.7, and 0.9, and the averaged results across these perturbation levels are presented in Table~\ref{tab:perturbation_avg_results}. The results show that PA-RNet consistently maintains superior forecasting performance under increasing textual noise, indicating its robustness against corrupted, irrelevant, or incomplete textual inputs. Additional comparisons with multimodal and foundation-model-based forecasting methods, together with the complete experimental results for the above evaluations, are provided in Appendix~\ref{app:additional comparison} and Appendix~\ref{app:complete_results}, respectively.

% Interestingly, we observe that several existing multimodal time-series forecasting models underperform their unimodal time-series-only counterparts on some datasets. This suggests that directly incorporating textual information does not always improve forecasting performance, especially when the textual modality contains noisy, weakly relevant, or misleading content. In such cases, the model may be distracted by unreliable textual cues, leading to degraded predictions.
Although PA-RNet achieves overall superior performance, we observe slight degradation on a few dataset-backbone combinations. This may be attributed to three main factors. First, for datasets such as Energy and Environment, the numerical modality already contains strong periodic or continuous temporal patterns, making the additional textual information less beneficial. In such cases, the perturbation-aware projection module and cross-modal attention modules may introduce extra optimization complexity. Second, for frequency-aware backbones such as FEDformer, the proposed spectral residual correction may partially overlap with the backbone's own frequency decomposition mechanism, potentially leading to over-smoothing of useful high-frequency variations. Third, on datasets with more complex event-driven semantics, such as Social Good, simple textual refinement may not fully distinguish useful semantic information from noisy descriptions, resulting in unstable gains under certain backbones. In addition, the inconsistent changes between MSE and MAE on some datasets, such as Traffic, suggest that PA-RNet may reduce average errors while still being affected by a few peak or abrupt-change points. These observations indicate that future work should further explore adaptive fusion strategies and backbone-specific integration mechanisms.

These observations highlight the importance of perturbation-aware design in multimodal forecasting. PA-RNet addresses this issue by combining numerical perturbation-aware projection module, textual spectral residual correction, and cross-modal attention. These components help suppress unstable perturbation components, preserve task-relevant semantic information, and align corrected textual representations with temporal dynamics, thereby improving prediction stability and robustness under noisy multimodal conditions.

\subsection{Further Analysis}
\paragraph{Ablation on Key Modules.}  
To assess the effectiveness of the proposed components, we conduct ablation studies on three key modules of PA-RNet: the cross-attention mechanism, the frequency-domain enhancement module, and the perturbation-aware projection module. We compare the full model with three variants: \textit{w/o Cross-Attention}, where cross-attention is replaced by simple concatenation; \textit{w/o spectral residual correction}, where frequency-domain modeling is removed; and \textit{w/o perturbation-aware projection module}, where the perturbation-aware projection module is discarded and the original textual embeddings are directly used.

All variants are trained under the same settings as the full model for a fair comparison. The results in Table~\ref{tab:ablation} show that removing any component leads to a noticeable drop in performance in most cases, indicating that these modules are complementary and jointly contribute to accurate and robust multimodal time series forecasting. 

\begin{figure*}[htbp]
    \centering
    \includegraphics[width=1.0\linewidth]{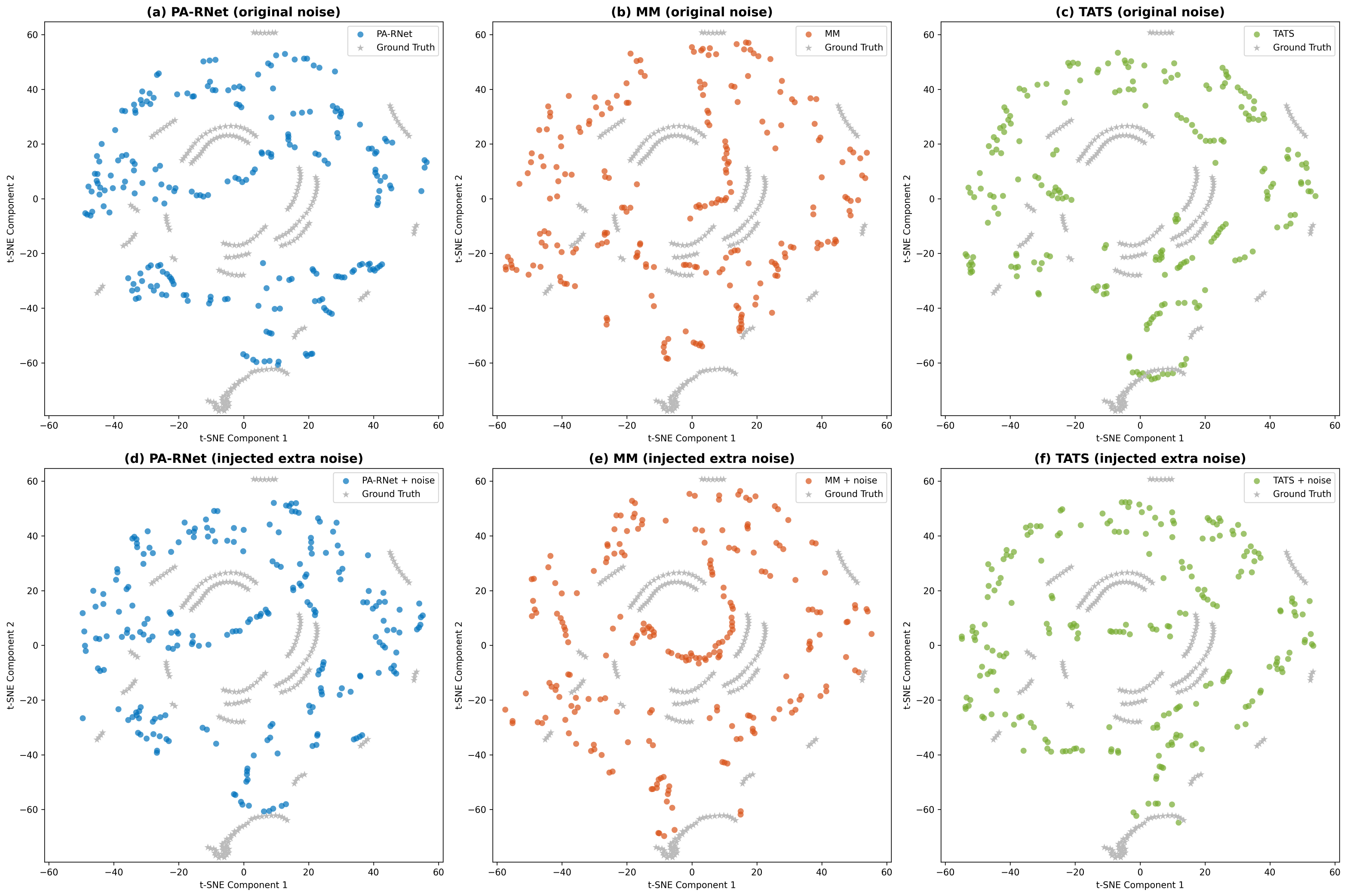}
    \caption{t-SNE visualization on the Health dataset before and after injecting extra textual noise.}
\label{fig:tsne_case}
\end{figure*}

\paragraph{Visualization Case Study.}
We provide a t-SNE visualization on the Health dataset to qualitatively assess the robustness of different multimodal forecasting frameworks. Autoformer is adopted as the backbone, and the prediction length is set to 48. As shown in Figure~\ref{fig:tsne_case}, PA-RNet preserves a more stable prediction distribution after textual noise injection, with its projected points remaining better aligned with the ground-truth distribution. In contrast, MM-TSFLib and TaTS show larger distributional deviations under noisy textual conditions. This visualization further confirms that PA-RNet is more effective in resisting textual perturbations and maintaining reliable forecasting representations.

% \paragraph{HyperParameter Setting.} Table~\ref{tab:hyperparams} summarizes the training configurations. For fair comparison, we follow official and original paper settings as baselines, with further tuning per dataset. To accommodate datasets with different temporal resolutions, we use adaptive batch sizes: 32 for monthly, 16 for weekly, and 8 for daily data. This improves memory efficiency and training stability by matching batch size to data granularity. Correspondingly, input configurations are set as follows: for monthly data, lookback 8, label 4, prediction 6; for weekly, 36, 18, 12; and for daily, 96, 48, 48.\label{app:hyperparameter_settings}
\paragraph{Hyperparameter Settings.}
Appendix~\ref{app:hyperparameter_settings} summarizes the main hyperparameter settings used for model training. We tune key parameters such as sequence length, learning rate, rank, prior weight, and module-specific learning rates, while keeping other training configurations fixed across experiments.
% Table~\ref{tab:hyperparams} summarizes the hyperparameter settings used in our training. We follow official implementations and original paper settings to ensure fair and consistent comparisons.
% To accommodate datasets with different temporal resolutions, we use adaptive batch sizes: 32 for monthly, 16 for weekly, and 8 for daily data. This improves memory efficiency and training stability by matching batch size to data granularity. Correspondingly, input configurations are set as follows: for monthly data, lookback 8, label 4, prediction 6; for weekly, 36, 18, 12; and for daily, 96, 48, 48.

\begin{figure}[ht]
  \centering
  \includegraphics[width=1.0\textwidth]{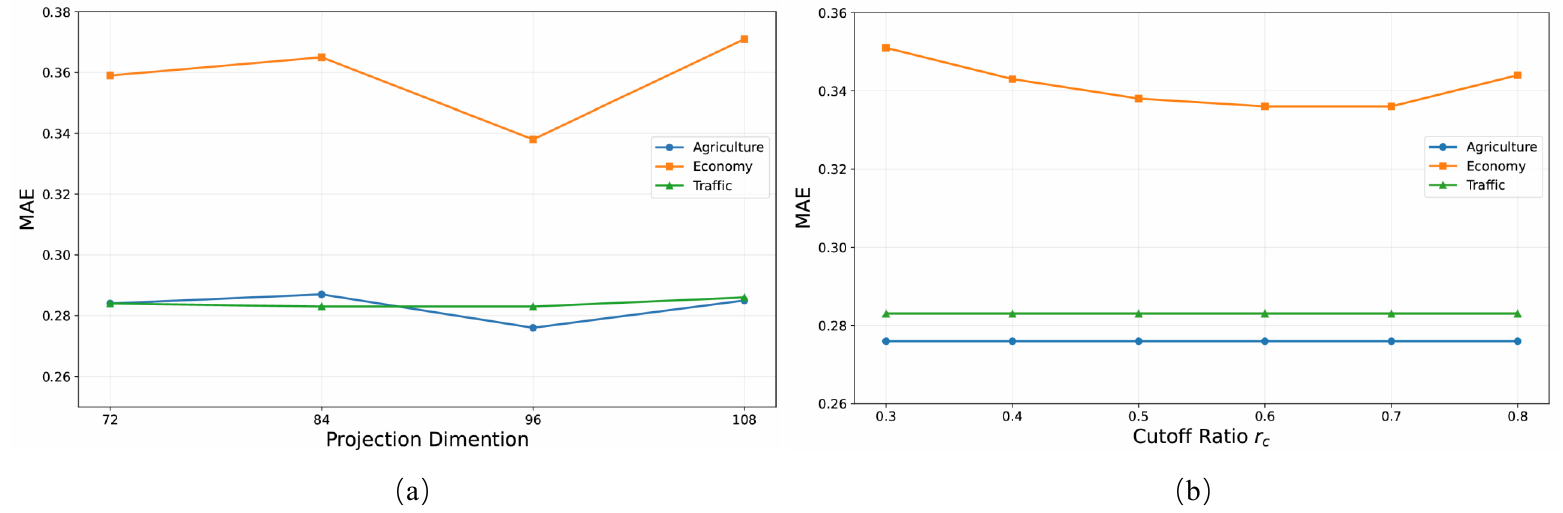}
\caption{Parameter sensitivity analysis of PA-RNet with Crossformer as the backbone forecasting model. 
(a) Projection dimension in the perturbation-aware numerical projection module. 
(b) Cutoff frequency ratio \(r_c\) in the textual spectral residual correction module.}
\label{fig:hyper_sensitivity}
\end{figure}

\paragraph{Parameter sensitivity analysis.}
To examine the influence of key hyperparameters, we conduct sensitivity analysis on PA-RNet with Crossformer as the backbone forecasting model. We study two hyperparameters: the projection dimension in the perturbation-aware numerical projection module and the cutoff frequency ratio \(r_c\) in the textual spectral residual correction module. The results are reported on the Agriculture, Economy, and Traffic datasets using MAE.
As shown in Figure~\ref{fig:hyper_sensitivity}(a), the projection dimension is varied among 72, 84, 96, and 108. The model obtains the best overall performance when the projection dimension is set to 96, indicating that this setting provides a suitable representation capacity for perturbation-related numerical components. Therefore, we use 96 as the default projection dimension.
Figure~\ref{fig:hyper_sensitivity}(b) reports the sensitivity of PA-RNet to the cutoff frequency ratio \(r_c\). The performance remains generally stable under different \(r_c\) values on the selected datasets, indicating that the textual spectral residual correction module is not highly sensitive to this hyperparameter within a reasonable range. Based on the overall performance and stability, we set \(r_c=0.5\) as the default value.

% We omit detailed sensitivity results for the cross-attention module, as it shares the same optimization strategy and exhibits similar robustness across learning rates.

\section{Conclusion and Future Work}
% \justifying % 使以下段落两端对齐
In this paper, we propose PA-RNet, a robust multimodal time-series forecasting framework for mitigating the negative effects of noisy textual information. By integrating perturbation-aware projection, spectral residual correction, and cross-modal attention, PA-RNet refines numerical and textual representations while aligning textual semantics with temporal dynamics. We also provide theoretical analysis to establish its Lipschitz continuity with respect to textual embeddings and show that spectral residual correction can reduce the expected prediction error under zero-mean textual perturbations. Experiments on nine multimodal time-series datasets demonstrate that PA-RNet maintains stable performance under both original textual conditions and simulated textual perturbations, confirming its effectiveness and robustness.
In future work, we plan to explore more adaptive multimodal fusion strategies, extend PA-RNet to broader temporal prediction tasks, and investigate its integration with large-scale pretrained models for more general and data-efficient multimodal forecasting.

\bibliographystyle{plainnat}
\bibliography{references}

%%%%%%%%%%%%%%%%%%%%%%%%%%%%%%%%%%%%%%%%%%%%%%%%%%%%%%%%%%%%
% \newpage
\appendix

\section{Theoretical Proofs of Robustness Analysis}
\label{app:theoretical_proofs}
To analyze the robustness of the textual branch, we assume that the textual embedding \(e_t\) can be decomposed into two orthogonal components: a task-relevant semantic signal component and a perturbation component:
\[
e_t
=
e_t^{\mathrm{signal}}
+
e_t^{\mathrm{noise}},
\]
where
\[
e_t^{\mathrm{signal}}
\perp
e_t^{\mathrm{noise}}.
\]
The perturbation component \(e_t^{\mathrm{noise}}\) is assumed to be an independent zero-mean noise term:
\[
\mathbb{E}\left[e_t^{\mathrm{noise}}\right]=0.
\]

Instead of directly subtracting the noise component, the proposed textual module learns a frequency-aware residual correction:
\[
\Psi_e(e_t)
=
\mathcal{F}^{-1}
\left(
M(\omega)\odot \mathcal{F}(e_tUV)
\right).
\]
Thus, the denoised textual embedding can be written as
\[
\widetilde{e}_t
=
e_t+\Psi_e(e_t).
\]
Ideally, the residual correction \(\Psi_e(e_t)\) approximates the negative perturbation component, i.e.,
\[
\Psi_e(e_t) \approx - e_t^{\mathrm{noise}}.
\]
Due to the approximation error of the low-rank spectral correction module, we write
\[
\Psi_e(e_t)
=
- e_t^{\mathrm{noise}}
+
\gamma_t,
\]
where \(\gamma_t\) denotes the residual perturbation caused by imperfect noise estimation. Therefore, the denoised textual embedding becomes
\[
\widetilde{e}_t
=
e_t+\Psi_e(e_t)
=
e_t^{\mathrm{signal}}
+
e_t^{\mathrm{noise}}
-
e_t^{\mathrm{noise}}
+
\gamma_t
=
e_t^{\mathrm{signal}}
+
\gamma_t.
\]
We assume that the residual perturbation is much smaller than the original noise component and remains zero-mean:
\[
\|\gamma_t\| \ll \|e_t^{\mathrm{noise}}\|,
\qquad
\mathbb{E}\left[\gamma_t\right]=0.
\]

Accordingly, the effective perturbation removed by the textual residual correction module is
\[
\eta_t
=
e_t-\widetilde{e}_t
=
e_t^{\mathrm{noise}}-\gamma_t.
\]
Since both \(e_t^{\mathrm{noise}}\) and \(\gamma_t\) are assumed to be zero-mean, we have
\[
\mathbb{E}\left[\eta_t\right]
=
\mathbb{E}\left[e_t^{\mathrm{noise}}\right]
-
\mathbb{E}\left[\gamma_t\right]
=
0.
\]
\begin{proposition}[Lipschitz Continuity]
\label{prop:lipschitz}
For a fixed observed time-series input \(x\), consider the proposed PA-RNet model
\[
f(x,e_t)
=
F\left(
x \mathbin{\|}
A\left(
\widetilde{x},
\widetilde{e}_t
\right)
\right),
\]
where
\[
\widetilde{x}
=
\mathrm{LN}
\left(
x-\sigma(\Gamma_x(x))\odot \Phi_x(x)
\right),
\]
and
\[
\widetilde{e}_t
=
e_t
+
\mathcal{F}^{-1}
\left(
M(\omega)\odot \mathcal{F}(e_tUV)
\right).
\]
Assume that the forecasting backbone \(F(\cdot)\), the cross-attention module
\(A(\cdot,\cdot)\), and the spectral residual correction module are Lipschitz
continuous on the considered input domain. Then \(f(x,e_t)\) is Lipschitz
continuous with respect to the textual embedding \(e_t\). That is, there exists
a constant \(L>0\) such that
\[
\forall e_t,e_t'\in \mathbb{R}^d,\qquad
\|f(x,e_t)-f(x,e_t')\|
\leq
L\|e_t-e_t'\|.
\]
\end{proposition}

\begin{proof}
For notational simplicity, define the spectral residual correction module as
\[
\Psi_e(e_t)
=
\mathcal{F}^{-1}
\left(
M(\omega)\odot \mathcal{F}(e_tUV)
\right).
\]
Then the corrected textual embedding can be written as
\[
\widetilde{e}_t
=
e_t+\Psi_e(e_t).
\]

\textbf{Step 1: Lipschitz continuity of the spectral residual correction module.}

The mapping \(e_t \mapsto e_tUV\) is a linear transformation. Therefore, for any
\(u,v\in\mathbb{R}^d\), we have
\[
\|uUV-vUV\|
=
\|(u-v)UV\|
\leq
\|U\|_2\|V\|_2\|u-v\|,
\]
where \(\|\cdot\|_2\) denotes the spectral norm.

The Fourier transform \(\mathcal{F}(\cdot)\) and inverse Fourier transform
\(\mathcal{F}^{-1}(\cdot)\) are bounded linear operators. Let their operator
norms be \(C_{\mathcal{F}}\) and \(C_{\mathcal{F}^{-1}}\), respectively.
Moreover, the frequency mask \(M(\omega)\) is a bounded element-wise multiplier.
Let
\[
\|M\|_{\infty}
=
\sup_{\omega}|M(\omega)|.
\]
For the low-pass mask used in our model, typically \(\|M\|_{\infty}\leq 1\).

Therefore, for any \(u,v\), we have
\begin{align*}
\|\Psi_e(u)-\Psi_e(v)\|
&=
\left\|
\mathcal{F}^{-1}
\left(
M(\omega)\odot
\mathcal{F}(uUV-vUV)
\right)
\right\| \\
&\leq
C_{\mathcal{F}^{-1}}
\|M\|_{\infty}
C_{\mathcal{F}}
\|U\|_2\|V\|_2
\|u-v\|.
\end{align*}
Thus, \(\Psi_e(\cdot)\) is Lipschitz continuous with constant
\[
L_{\Psi}
=
C_{\mathcal{F}^{-1}}
\|M\|_{\infty}
C_{\mathcal{F}}
\|U\|_2\|V\|_2.
\]

Since
\[
\widetilde{e}
=
e+\Psi_e(e),
\]
we further obtain
\begin{align*}
\|\widetilde{u}-\widetilde{v}\|
&=
\|(u+\Psi_e(u))-(v+\Psi_e(v))\| \\
&\leq
\|u-v\|+\|\Psi_e(u)-\Psi_e(v)\| \\
&\leq
(1+L_{\Psi})\|u-v\|.
\end{align*}
Hence, the corrected textual embedding mapping
\[
e \mapsto \widetilde{e}
\]
is Lipschitz continuous with constant \(1+L_{\Psi}\).

\textbf{Step 2: Lipschitz continuity of the cross-attention module.}

The cross-attention module takes \(\widetilde{x}\) as the query-side input and
\(\widetilde{e}_t\) as the key-value-side input. Its general form can be written as
\[
\mathrm{CrossAttn}(Q,K,V)
=
\mathrm{Softmax}
\left(
\frac{QK^\top}{\sqrt{d}}
\right)V,
\]
where
\[
Q=\widetilde{x}W_Q,\qquad
K=\widetilde{e}_tW_K,\qquad
V=\widetilde{e}_tW_V.
\]
For fixed \(x\), the representation \(\widetilde{x}\) is also fixed with respect
to \(e_t\). The mappings from \(\widetilde{e}_t\) to \(K\) and \(V\) are linear
and thus Lipschitz continuous. The softmax function is Lipschitz continuous on
bounded domains, and the product operations involved in attention are also
Lipschitz continuous on bounded domains. Therefore, the cross-attention module
is Lipschitz continuous with respect to \(\widetilde{e}_t\). That is, there exists
a constant \(L_A>0\) such that
\[
\left\|
A(\widetilde{x},\widetilde{u})
-
A(\widetilde{x},\widetilde{v})
\right\|
\leq
L_A
\|\widetilde{u}-\widetilde{v}\|.
\]

Combining this inequality with the result from Step 1, we have
\[
\left\|
A(\widetilde{x},\widetilde{u})
-
A(\widetilde{x},\widetilde{v})
\right\|
\leq
L_A(1+L_{\Psi})\|u-v\|.
\]

\textbf{Step 3: Lipschitz continuity of the forecasting backbone.}

The forecasting backbone \(F(\cdot)\) can be instantiated by Transformer, GRU,
LSTM, or other neural forecasting architectures. These models are composed of
linear transformations, activation functions, normalization operations, and
attention or recurrent units. Under bounded parameters and bounded input
domains, each component is Lipschitz continuous. Therefore, the entire backbone
\(F(\cdot)\) is Lipschitz continuous. Let its Lipschitz constant be \(L_F>0\).
Then for any two inputs \(z_1,z_2\), we have
\[
\|F(x\mathbin{\|}z_1)-F(x\mathbin{\|}z_2)\|
\leq
L_F\|z_1-z_2\|.
\]
Since the time-series input \(x\) is fixed, the concatenation operation does not
change the difference induced by the textual branch:
\[
\|(x\mathbin{\|}z_1)-(x\mathbin{\|}z_2)\|
=
\|z_1-z_2\|.
\]

\noindent\textbf{Final bound.}
Combining the above results, for any \(e_t,e_t'\in\mathbb{R}^d\), we obtain
\begin{align*}
\|f(x,e_t)-f(x,e_t')\|
&=
\left\|
F\left(
x\mathbin{\|}
A(\widetilde{x},\widetilde{e}_t)
\right)
-
F\left(
x\mathbin{\|}
A(\widetilde{x},\widetilde{e}_t')
\right)
\right\| \\
&\leq
L_F
\left\|
A(\widetilde{x},\widetilde{e}_t)
-
A(\widetilde{x},\widetilde{e}_t')
\right\| \\
&\leq
L_F L_A
\|\widetilde{e}_t-\widetilde{e}_t'\| \\
&\leq
L_F L_A(1+L_{\Psi})
\|e_t-e_t'\|.
\end{align*}
Therefore, \(f(x,e_t)\) is Lipschitz continuous with respect to the textual
embedding \(e_t\), with Lipschitz constant
\[
L
=
L_F L_A(1+L_{\Psi}),
\]
where
\[
L_{\Psi}
=
C_{\mathcal{F}^{-1}}
\|M\|_{\infty}
C_{\mathcal{F}}
\|U\|_2\|V\|_2.
\]
\end{proof}

\begin{proposition}[Expected Error Reduction via Spectral Residual Correction]
\label{prop:error_reduction}
Let \(f(x,e_t)\) be the prediction function of PA-RNet, where the corrected textual embedding is defined as
\[
\widetilde{e}_t
=
e_t+\Psi_e(e_t),
\]
with
\[
\Psi_e(e_t)
=
\mathcal{F}^{-1}
\left(
M(\omega)\odot \mathcal{F}(e_tUV)
\right).
\]
Assume that the textual embedding can be decomposed as
\[
e_t
=
e_t^{\mathrm{signal}}
+
e_t^{\mathrm{noise}},
\]
where \(e_t^{\mathrm{noise}}\) is an independent zero-mean perturbation:
\[
\mathbb{E}\left[e_t^{\mathrm{noise}}\right]=0.
\]
The spectral residual correction module is expected to learn a compensatory residual that cancels the noise component:
\[
\Psi_e(e_t)
=
-
e_t^{\mathrm{noise}}
+
\gamma_t,
\]
where \(\gamma_t\) is a residual perturbation satisfying
\[
\mathbb{E}[\gamma_t]=0,
\qquad
\mathbb{E}\left[\|\gamma_t\|^2\right]
<
\mathbb{E}\left[\|e_t^{\mathrm{noise}}\|^2\right].
\]
If \(f\) is Lipschitz continuous with respect to the textual embedding and the perturbations are output-unbiased, then the corrected embedding reduces the expected prediction error:
\[
\mathbb{E}\left[
\mathcal{L}(f(x,\widetilde{e}_t),y_t)
\right]
<
\mathbb{E}\left[
\mathcal{L}(f(x,e_t),y_t)
\right],
\]
where
\[
\mathcal{L}(f(x,e_t),y_t)
=
\|f(x,e_t)-y_t\|^2.
\]
\end{proposition}

\begin{proof}
According to the decomposition of the textual embedding, we have
\[
e_t
=
e_t^{\mathrm{signal}}
+
e_t^{\mathrm{noise}}.
\]
The proposed spectral residual correction module produces
\[
\Psi_e(e_t)
=
-
e_t^{\mathrm{noise}}
+
\gamma_t.
\]
Therefore, the corrected textual embedding becomes
\begin{align*}
\widetilde{e}_t
&=
e_t+\Psi_e(e_t) \\
&=
e_t^{\mathrm{signal}}
+
e_t^{\mathrm{noise}}
-
e_t^{\mathrm{noise}}
+
\gamma_t \\
&=
e_t^{\mathrm{signal}}
+
\gamma_t.
\end{align*}
Thus, the original textual embedding contains the full noise perturbation \(e_t^{\mathrm{noise}}\), while the corrected embedding only contains the residual perturbation \(\gamma_t\).

Let
\[
m
\coloneqq
f(x,e_t^{\mathrm{signal}})
\]
denote the model output under the clean textual signal. Define the model outputs under the original noisy embedding and the corrected embedding as
\[
Z_{\mathrm{noise}}
\coloneqq
f(x,e_t)
=
f(x,e_t^{\mathrm{signal}}+e_t^{\mathrm{noise}}),
\]
and
\[
Z_{\mathrm{corr}}
\coloneqq
f(x,\widetilde{e}_t)
=
f(x,e_t^{\mathrm{signal}}+\gamma_t).
\]

Since both \(e_t^{\mathrm{noise}}\) and \(\gamma_t\) are assumed to be zero-mean perturbations, and the model is locally smooth, we assume that the perturbations are output-unbiased:
\[
\mathbb{E}
\left[
Z_{\mathrm{noise}}
\mid x,e_t^{\mathrm{signal}}
\right]
=
m,
\]
and
\[
\mathbb{E}
\left[
Z_{\mathrm{corr}}
\mid x,e_t^{\mathrm{signal}}
\right]
=
m.
\]

Using the bias-variance decomposition of the squared loss, we have
\[
\mathbb{E}
\left[
\|Z_{\mathrm{noise}}-y_t\|^2
\right]
=
\|m-y_t\|^2
+
\mathbb{E}
\left[
\|Z_{\mathrm{noise}}-m\|^2
\right],
\]
and
\[
\mathbb{E}
\left[
\|Z_{\mathrm{corr}}-y_t\|^2
\right]
=
\|m-y_t\|^2
+
\mathbb{E}
\left[
\|Z_{\mathrm{corr}}-m\|^2
\right].
\]

The first term \(\|m-y_t\|^2\) is the same for both cases. Therefore, the difference between the two expected losses depends on the variance term induced by the perturbation.

Since \(f\) is Lipschitz continuous with respect to the textual embedding, there exists a constant \(L_f>0\) such that
\[
\|f(x,u)-f(x,v)\|
\leq
L_f\|u-v\|.
\]
Applying this property to the corrected embedding gives
\begin{align*}
\|Z_{\mathrm{corr}}-m\|
&=
\left\|
f(x,e_t^{\mathrm{signal}}+\gamma_t)
-
f(x,e_t^{\mathrm{signal}})
\right\| \\
&\leq
L_f\|\gamma_t\|.
\end{align*}
Thus,
\[
\mathbb{E}
\left[
\|Z_{\mathrm{corr}}-m\|^2
\right]
\leq
L_f^2
\mathbb{E}
\left[
\|\gamma_t\|^2
\right].
\]

Similarly, for the original noisy embedding, we have
\[
\|Z_{\mathrm{noise}}-m\|
=
\left\|
f(x,e_t^{\mathrm{signal}}+e_t^{\mathrm{noise}})
-
f(x,e_t^{\mathrm{signal}})
\right\|.
\]
Since the spectral residual correction suppresses high-frequency perturbations and preserves low-frequency semantic components, the remaining perturbation \(\gamma_t\) has smaller energy than the original noise:
\[
\mathbb{E}
\left[
\|\gamma_t\|^2
\right]
<
\mathbb{E}
\left[
\|e_t^{\mathrm{noise}}\|^2
\right].
\]
Under the output-unbiased and variance-reduction assumptions, this implies
\[
\mathbb{E}
\left[
\|Z_{\mathrm{corr}}-m\|^2
\right]
<
\mathbb{E}
\left[
\|Z_{\mathrm{noise}}-m\|^2
\right].
\]

Consequently,
\begin{align*}
\mathbb{E}
\left[
\mathcal{L}(f(x,\widetilde{e}_t),y_t)
\right]
&=
\|m-y_t\|^2
+
\mathbb{E}
\left[
\|Z_{\mathrm{corr}}-m\|^2
\right] \\
&<
\|m-y_t\|^2
+
\mathbb{E}
\left[
\|Z_{\mathrm{noise}}-m\|^2
\right] \\
&=
\mathbb{E}
\left[
\mathcal{L}(f(x,e_t),y_t)
\right].
\end{align*}
Therefore, the proposed spectral residual correction mechanism reduces the expected prediction error under noisy textual perturbations.
\end{proof}

\section{Additional Comparison with Multimodal and Foundation-Model-Based Forecasting Methods}
\label{app:additional comparison}

\begin{table}[htbp]
\centering
\tiny
\caption{Comparison with multimodal and foundation-model-based forecasting methods. The best results are highlighted in bold.}
\resizebox{\textwidth}{!}{
\begin{tabular}{lcccccc}

\toprule
\multirow{2}{*}{Dataset} 
& \multicolumn{2}{c}{TimeVLM} 
& \multicolumn{2}{c}{TimeCMA} 
& \multicolumn{2}{c}{PA-RNet (iTransformer)} \\
\cmidrule(lr){2-3} \cmidrule(lr){4-5} \cmidrule(lr){6-7}
& MSE & MAE & MSE & MAE & MSE & MAE \\
\midrule
Agriculture 
& 0.302 & 0.360 
& 0.116 & 0.219 
& \textbf{0.094} & \textbf{0.204} \\

Climate 
& 1.186 & 0.886 
& 1.518 & 1.001 
& \textbf{1.016} & \textbf{0.807} \\

Economy 
& 0.061 & 0.196 
& 0.045 & 0.173 
& \textbf{0.010} & \textbf{0.084} \\

Energy 
& 0.344 & 0.457 
& 0.414 & 0.497 
& \textbf{0.279} & \textbf{0.398} \\

Environment 
& 0.546 & 0.608 
& 0.353 & 0.420 
& \textbf{0.259} & \textbf{0.370} \\

Health 
& 1.907 & 1.019 
& 2.694 & 1.205 
& \textbf{1.339} & \textbf{0.748} \\

Security 
& 124.522 & 5.647 
& 130.984 & 6.192 
& \textbf{107.554} & \textbf{4.816} \\

Social Good 
& \textbf{1.033} & \textbf{0.544} 
& 1.180 & 0.581 
& 1.136 & 0.634 \\

Traffic 
& 0.205 & 0.309 
& 0.350 & 0.456 
& \textbf{0.178} & \textbf{0.221} \\
\bottomrule
\end{tabular}
}
\vspace{4pt}

\label{tab:comparison_timevlm_timecma}
\end{table}

\section{Hyperparameter Settings}
\label{app:hyperparameter_settings}
\begin{table}[ht]
\centering
\footnotesize
\caption{Hyperparameter Settings for Model Training}
\begin{tabular}{llc}
\toprule
\textbf{Hyperparameter} & \textbf{Description} & \textbf{Value or Choices} \\
\midrule
Batch size & Training batch size & 32 \\
Seq\_len & Input sequence length & \{8, 36, 96\}\\
Label\_len & Label sequence length & \{4, 18, 48\}\\
Criterion & Loss function & Mean Square Error (MSE) \\
Model learning rate & Learning rate for the time series backbone model & \{0.01, 0.001, 0.0001, 0.00001\} \\
Rank & Rank of the linear layer before FFT & \{8, 16, 32, 64\} \\

% Maximum length & Maximum length for relative positional encoding & \{8, 32, 72\} \\
Module learning rate & spectral residual, cross-attention, perturbation modules & $\{10^{-3}, 10^{-4}, 10^{-5}, 10^{-6}\}$ \\
Cross-attention heads & Number of attention heads in cross-attention & 8 \\
Prior weight & Weight for prior combination & \{0,0.1,0.2,0.3,0.5\} \\
Training epochs & Total number of training epochs & 50 \\
Patience & Early stopping patience & 20 \\
Text embedding dimension & Dimension of textual embeddings & 12 \\
Pooling type & Embedding pooling method & avg \\
Initialization method & Weight initialization method & normal \\
Dropout rate & Dropout probability & 0.1 \\
Use normalization & Whether to use normalization & TRUE \\
\bottomrule
\end{tabular}
\vspace{4pt}

\label{tab:hyperparams}

\end{table}

\section{Complete Results under Original and Perturbed Textual Conditions}
\label{app:complete_results}

\begin{table*}[t]
\centering
\tiny
\setlength{\tabcolsep}{1.6pt}
\renewcommand{\arraystretch}{0.92}
\caption{Detailed forecasting results on Agriculture, Climate, and Economy under the original noisy textual conditions. Lower MSE/MAE indicates better performance.}
\label{tab:complete_original_noisy_results_three_datasets}
\resizebox{\textwidth}{!}{
\begin{tabular}{llc|cc|cc|cc|cc|cc|cc|cc|cc|cc}
\toprule
\multirow{2}{*}{Dataset} & \multirow{2}{*}{Method} & \multirow{2}{*}{$T$}
& \multicolumn{2}{c|}{Autoformer}
& \multicolumn{2}{c|}{Crossformer}
& \multicolumn{2}{c|}{DLinear}
& \multicolumn{2}{c|}{FEDformer}
& \multicolumn{2}{c|}{FiLM}
& \multicolumn{2}{c|}{Informer}
& \multicolumn{2}{c|}{iTransformer}
& \multicolumn{2}{c|}{PatchTST}
& \multicolumn{2}{c}{Transformer} \\
\cmidrule(lr){4-5} \cmidrule(lr){6-7} \cmidrule(lr){8-9} \cmidrule(lr){10-11} \cmidrule(lr){12-13} \cmidrule(lr){14-15} \cmidrule(lr){16-17} \cmidrule(lr){18-19} \cmidrule(lr){20-21}
& & & MSE & MAE & MSE & MAE & MSE & MAE & MSE & MAE & MSE & MAE & MSE & MAE & MSE & MAE & MSE & MAE & MSE & MAE \\
\midrule

\multirow{20}{*}{Agriculture}
& \multirow{5}{*}{Uni-modal}
& 6   & 0.078 & 0.205 & 0.173 & 0.303 & 0.156 & 0.302 & 0.074 & 0.195 & 0.066 & 0.160 & 0.272 & 0.384 & 0.067 & 0.180 & 0.059 & 0.160 & 0.199 & 0.315 \\
& & 8   & 0.091 & 0.217 & 0.266 & 0.373 & 0.173 & 0.320 & 0.090 & 0.219 & 0.084 & 0.183 & 0.315 & 0.403 & 0.079 & 0.188 & 0.079 & 0.183 & 0.208 & 0.311 \\
& & 10  & 0.110 & 0.248 & 0.362 & 0.450 & 0.206 & 0.343 & 0.105 & 0.247 & 0.108 & 0.209 & 0.490 & 0.530 & 0.102 & 0.219 & 0.101 & 0.206 & 0.305 & 0.390 \\
& & 12  & 0.134 & 0.269 & 0.436 & 0.495 & 0.247 & 0.371 & 0.132 & 0.273 & 0.142 & 0.241 & 0.578 & 0.567 & 0.134 & 0.247 & 0.131 & 0.238 & 0.419 & 0.451 \\
& & Avg & 0.103 & 0.235 & 0.309 & 0.405 & 0.195 & 0.334 & 0.100 & 0.234 & 0.100 & 0.198 & 0.414 & 0.471 & 0.096 & 0.209 & 0.093 & 0.197 & 0.283 & 0.367 \\
\cmidrule(lr){2-21}

& \multirow{5}{*}{MM-TSFLib}
& 6   & 0.076 & 0.198 & 0.138 & 0.258 & 0.098 & 0.214 & 0.073 & 0.194 & 0.067 & 0.161 & 0.203 & 0.319 & 0.063 & 0.173 & 0.061 & 0.163 & 0.164 & 0.281 \\
\bottomrule
\end{tabular}
}
\end{table*}

\begin{table*}[t]
\centering
\tiny
\setlength{\tabcolsep}{1.6pt}
\renewcommand{\arraystretch}{0.92}
\caption{Detailed forecasting results on Energy, Environment, and Health under the original noisy textual conditions. Lower MSE/MAE indicates better performance.}
\label{tab:complete_original_noisy_results_energy_environment_health}
\resizebox{\textwidth}{!}{
\begin{tabular}{llc|cc|cc|cc|cc|cc|cc|cc|cc|cc}
\toprule
\multirow{2}{*}{Dataset} & \multirow{2}{*}{Method} & \multirow{2}{*}{$T$}
& \multicolumn{2}{c|}{Autoformer}
& \multicolumn{2}{c|}{Crossformer}
& \multicolumn{2}{c|}{DLinear}
& \multicolumn{2}{c|}{FEDformer}
& \multicolumn{2}{c|}{FiLM}
& \multicolumn{2}{c|}{Informer}
& \multicolumn{2}{c|}{iTransformer}
& \multicolumn{2}{c|}{PatchTST}
& \multicolumn{2}{c}{Transformer} \\
\cmidrule(lr){4-5} \cmidrule(lr){6-7} \cmidrule(lr){8-9} \cmidrule(lr){10-11} \cmidrule(lr){12-13} \cmidrule(lr){14-15} \cmidrule(lr){16-17} \cmidrule(lr){18-19} \cmidrule(lr){20-21}
& & & MSE & MAE & MSE & MAE & MSE & MAE & MSE & MAE & MSE & MAE & MSE & MAE & MSE & MAE & MSE & MAE & MSE & MAE \\
\midrule

\multirow{20}{*}{Energy}
& \multirow{5}{*}{Uni-modal}
& 12  & 0.177 & 0.321 & 0.142 & 0.273 & 0.104 & 0.230 & 0.097 & 0.218 & 0.101 & 0.225 & 0.200 & 0.314 & 0.125 & 0.263 & 0.146 & 0.283 & 0.179 & 0.297 \\
& & 24  & 0.334 & 0.455 & 0.297 & 0.413 & 0.218 & 0.336 & 0.180 & 0.307 & 0.217 & 0.340 & 0.320 & 0.442 & 0.253 & 0.381 & 0.254 & 0.382 & 0.277 & 0.402 \\
& & 36  & 0.403 & 0.503 & 0.403 & 0.487 & 0.298 & 0.395 & 0.253 & 0.372 & 0.310 & 0.412 & 0.476 & 0.543 & 0.326 & 0.438 & 0.352 & 0.460 & 0.493 & 0.537 \\
& & 48  & 0.513 & 0.567 & 0.481 & 0.542 & 0.406 & 0.470 & 0.366 & 0.470 & 0.415 & 0.491 & 0.551 & 0.605 & 0.447 & 0.530 & 0.449 & 0.530 & 0.528 & 0.589 \\
& & Avg & 0.357 & 0.461 & 0.331 & 0.429 & 0.257 & 0.358 & 0.224 & 0.342 & 0.261 & 0.367 & 0.387 & 0.476 & 0.288 & 0.403 & 0.300 & 0.414 & 0.369 & 0.456 \\
\cmidrule(lr){2-21}

& \multirow{5}{*}{MM-TSFLib}
& 12  & 0.168 & 0.315 & 0.131 & 0.257 & 0.103 & 0.229 & 0.090 & 0.212 & 0.099 & 0.221 & 0.209 & 0.342 & 0.135 & 0.278 & 0.135 & 0.272 & 0.220 & 0.344 \\
\bottomrule
\end{tabular}
}
\end{table*}

\begin{table*}[t]
\centering
\tiny
\setlength{\tabcolsep}{1.6pt}
\renewcommand{\arraystretch}{0.92}
\caption{Detailed forecasting results on Security, Social Good, and Traffic under the original noisy textual conditions. Lower MSE/MAE indicates better performance.}
\label{tab:complete_original_noisy_results_security_social_traffic}
\resizebox{\textwidth}{!}{
\begin{tabular}{llc|cc|cc|cc|cc|cc|cc|cc|cc|cc}
\toprule
\multirow{2}{*}{Dataset} & \multirow{2}{*}{Method} & \multirow{2}{*}{$T$}
& \multicolumn{2}{c|}{Autoformer}
& \multicolumn{2}{c|}{Crossformer}
& \multicolumn{2}{c|}{DLinear}
& \multicolumn{2}{c|}{FEDformer}
& \multicolumn{2}{c|}{FiLM}
& \multicolumn{2}{c|}{Informer}
& \multicolumn{2}{c|}{iTransformer}
& \multicolumn{2}{c|}{PatchTST}
& \multicolumn{2}{c}{Transformer} \\
\cmidrule(lr){4-5} \cmidrule(lr){6-7} \cmidrule(lr){8-9} \cmidrule(lr){10-11} \cmidrule(lr){12-13} \cmidrule(lr){14-15} \cmidrule(lr){16-17} \cmidrule(lr){18-19} \cmidrule(lr){20-21}
& & & MSE & MAE & MSE & MAE & MSE & MAE & MSE & MAE & MSE & MAE & MSE & MAE & MSE & MAE & MSE & MAE & MSE & MAE \\
\midrule

\multirow{20}{*}{Security}
& \multirow{5}{*}{Uni-modal}
& 6   & 125.784 & 5.469 & 126.449 & 6.161 & 104.875 & 4.960 & 144.717 & 5.981 & 119.227 & 5.555 & 126.891 & 6.277 & 104.867 & 4.791 & 126.968 & 6.145 & 123.389 & 5.983 \\
& & 8   & 125.874 & 5.406 & 127.465 & 6.308 & 108.280 & 5.126 & 132.596 & 5.584 & 130.614 & 6.380 & 129.400 & 6.504 & 106.625 & 4.888 & 125.737 & 5.985 & 126.223 & 6.239 \\
& & 10  & 126.364 & 5.435 & 125.520 & 6.214 & 111.262 & 5.324 & 128.839 & 5.569 & 133.139 & 6.528 & 132.062 & 6.720 & 108.510 & 4.958 & 124.925 & 5.909 & 127.792 & 6.381 \\
& & 12  & 122.761 & 5.364 & 126.094 & 6.292 & 113.018 & 5.374 & 126.916 & 5.537 & 134.097 & 6.572 & 133.043 & 6.809 & 112.233 & 5.167 & 126.309 & 6.016 & 128.924 & 6.481 \\
& & Avg & 125.196 & 5.419 & 126.382 & 6.244 & 109.359 & 5.196 & 133.267 & 5.668 & 129.269 & 6.259 & 130.349 & 6.578 & 108.059 & 4.951 & 125.985 & 6.014 & 126.582 & 6.271 \\
\cmidrule(lr){2-21}

& \multirow{5}{*}{MM-TSFLib}
& 6   & 135.978 & 5.781 & 127.760 & 6.294 & 103.672 & 4.689 & 141.976 & 5.859 & 113.719 & 5.244 & 127.575 & 6.324 & 104.461 & 4.675 & 126.347 & 6.049 & 125.458 & 6.099 \\
\bottomrule
\end{tabular}
}
\end{table*}

\begin{table*}[t]
\centering
\tiny
\setlength{\tabcolsep}{1.6pt}
\renewcommand{\arraystretch}{0.92}
\caption{Detailed forecasting results on Agriculture, Climate, and Economy under textual perturbation ratio $\rho=0.3$. Lower MSE/MAE indicates better performance.}
\label{tab:complete_perturbation_03_results_agriculture_climate_economy}
\resizebox{\textwidth}{!}{
\begin{tabular}{llc|cc|cc|cc|cc|cc|cc|cc|cc|cc}
\toprule
\multirow{2}{*}{Dataset} & \multirow{2}{*}{Method} & \multirow{2}{*}{$T$}
& \multicolumn{2}{c|}{Autoformer}
& \multicolumn{2}{c|}{Crossformer}
& \multicolumn{2}{c|}{DLinear}
& \multicolumn{2}{c|}{FEDformer}
& \multicolumn{2}{c|}{FiLM}
& \multicolumn{2}{c|}{Informer}
& \multicolumn{2}{c|}{iTransformer}
& \multicolumn{2}{c|}{PatchTST}
& \multicolumn{2}{c}{Transformer} \\
\cmidrule(lr){4-5} \cmidrule(lr){6-7} \cmidrule(lr){8-9} \cmidrule(lr){10-11} \cmidrule(lr){12-13} \cmidrule(lr){14-15} \cmidrule(lr){16-17} \cmidrule(lr){18-19} \cmidrule(lr){20-21}
& & & MSE & MAE & MSE & MAE & MSE & MAE & MSE & MAE & MSE & MAE & MSE & MAE & MSE & MAE & MSE & MAE & MSE & MAE \\
\midrule

\multirow{20}{*}{Agriculture}
& \multirow{5}{*}{Uni-modal}
& 6   & 0.078 & 0.205 & 0.173 & 0.303 & 0.156 & 0.302 & 0.074 & 0.195 & 0.066 & 0.160 & 0.272 & 0.384 & 0.067 & 0.180 & 0.059 & 0.160 & 0.199 & 0.315 \\
& & 8   & 0.091 & 0.217 & 0.266 & 0.373 & 0.173 & 0.320 & 0.090 & 0.219 & 0.084 & 0.183 & 0.315 & 0.403 & 0.079 & 0.188 & 0.079 & 0.183 & 0.208 & 0.311 \\
& & 10  & 0.110 & 0.248 & 0.362 & 0.450 & 0.206 & 0.343 & 0.105 & 0.247 & 0.108 & 0.209 & 0.490 & 0.530 & 0.102 & 0.219 & 0.101 & 0.206 & 0.305 & 0.390 \\
& & 12  & 0.134 & 0.269 & 0.436 & 0.495 & 0.247 & 0.371 & 0.132 & 0.273 & 0.142 & 0.241 & 0.578 & 0.567 & 0.134 & 0.247 & 0.131 & 0.238 & 0.419 & 0.451 \\
& & Avg & 0.103 & 0.235 & 0.309 & 0.405 & 0.195 & 0.334 & 0.100 & 0.234 & 0.100 & 0.198 & 0.414 & 0.471 & 0.096 & 0.209 & 0.093 & 0.197 & 0.283 & 0.367 \\
\cmidrule(lr){2-21}

& \multirow{5}{*}{MM-TSFLib}
& 6   & 0.076 & 0.199 & 0.138 & 0.260 & 0.094 & 0.211 & 0.073 & 0.194 & 0.067 & 0.161 & 0.211 & 0.328 & 0.072 & 0.200 & 0.061 & 0.163 & 0.158 & 0.276 \\
\bottomrule
\end{tabular}
}
\end{table*}

\begin{table*}[t]
\centering
\tiny
\setlength{\tabcolsep}{1.6pt}
\renewcommand{\arraystretch}{0.92}
\caption{Detailed forecasting results on Health, Security, and Traffic under textual perturbation ratio $\rho=0.3$. Lower MSE/MAE indicates better performance.}
\label{tab:complete_perturbation_03_results_health_security_traffic}
\resizebox{\textwidth}{!}{
\begin{tabular}{llc|cc|cc|cc|cc|cc|cc|cc|cc|cc}
\toprule
\multirow{2}{*}{Dataset} & \multirow{2}{*}{Method} & \multirow{2}{*}{$T$}
& \multicolumn{2}{c|}{Autoformer}
& \multicolumn{2}{c|}{Crossformer}
& \multicolumn{2}{c|}{DLinear}
& \multicolumn{2}{c|}{FEDformer}
& \multicolumn{2}{c|}{FiLM}
& \multicolumn{2}{c|}{Informer}
& \multicolumn{2}{c|}{iTransformer}
& \multicolumn{2}{c|}{PatchTST}
& \multicolumn{2}{c}{Transformer} \\
\cmidrule(lr){4-5} \cmidrule(lr){6-7} \cmidrule(lr){8-9} \cmidrule(lr){10-11} \cmidrule(lr){12-13} \cmidrule(lr){14-15} \cmidrule(lr){16-17} \cmidrule(lr){18-19} \cmidrule(lr){20-21}
& & & MSE & MAE & MSE & MAE & MSE & MAE & MSE & MAE & MSE & MAE & MSE & MAE & MSE & MAE & MSE & MAE & MSE & MAE \\
\midrule

\multirow{20}{*}{Health}
& \multirow{5}{*}{Uni-modal}
& 12  & 1.512 & 0.919 & 1.216 & 0.748 & 1.579 & 0.855 & 1.131 & 0.787 & 1.244 & 0.749 & 1.250 & 0.765 & 1.061 & 0.626 & 0.962 & 0.624 & 1.358 & 0.803 \\
& & 24  & 1.875 & 1.025 & 1.434 & 0.798 & 1.656 & 0.879 & 1.549 & 0.964 & 1.525 & 0.832 & 1.642 & 0.819 & 1.811 & 0.784 & 1.578 & 0.758 & 1.393 & 0.761 \\
& & 36  & 1.971 & 1.040 & 1.450 & 0.781 & 1.692 & 0.892 & 1.622 & 0.982 & 1.692 & 0.891 & 1.534 & 0.824 & 2.187 & 0.885 & 1.908 & 0.857 & 1.555 & 0.793 \\
& & 48  & 1.881 & 0.987 & 1.591 & 0.834 & 1.731 & 0.913 & 1.647 & 0.980 & 1.790 & 0.931 & 1.662 & 0.865 & 2.240 & 0.928 & 2.131 & 0.911 & 1.533 & 0.835 \\
& & Avg & 1.810 & 0.993 & 1.423 & 0.790 & 1.664 & 0.885 & 1.487 & 0.928 & 1.563 & 0.851 & 1.522 & 0.818 & 1.825 & 0.806 & 1.645 & 0.788 & 1.460 & 0.798 \\
\cmidrule(lr){2-21}

& \multirow{5}{*}{MM-TSFLib}
& 12  & 1.533 & 0.921 & 1.204 & 0.738 & 1.482 & 0.840 & 1.154 & 0.792 & 1.185 & 0.729 & 1.146 & 0.731 & 1.010 & 0.625 & 0.879 & 0.602 & 1.199 & 0.769 \\
\bottomrule
\end{tabular}
}
\end{table*}

\begin{table*}[t]
\centering
\tiny
\setlength{\tabcolsep}{1.6pt}
\renewcommand{\arraystretch}{0.92}
\caption{Detailed forecasting results on Agriculture, Climate, and Economy under textual perturbation ratio $\rho=0.5$. Lower MSE/MAE indicates better performance.}
\label{tab:complete_perturbation_05_results_agriculture_climate_economy}
\resizebox{\textwidth}{!}{
\begin{tabular}{llc|cc|cc|cc|cc|cc|cc|cc|cc|cc}
\toprule
\multirow{2}{*}{Dataset} & \multirow{2}{*}{Method} & \multirow{2}{*}{$T$}
& \multicolumn{2}{c|}{Autoformer}
& \multicolumn{2}{c|}{Crossformer}
& \multicolumn{2}{c|}{DLinear}
& \multicolumn{2}{c|}{FEDformer}
& \multicolumn{2}{c|}{FiLM}
& \multicolumn{2}{c|}{Informer}
& \multicolumn{2}{c|}{iTransformer}
& \multicolumn{2}{c|}{PatchTST}
& \multicolumn{2}{c}{Transformer} \\
\cmidrule(lr){4-5} \cmidrule(lr){6-7} \cmidrule(lr){8-9} \cmidrule(lr){10-11} \cmidrule(lr){12-13} \cmidrule(lr){14-15} \cmidrule(lr){16-17} \cmidrule(lr){18-19} \cmidrule(lr){20-21}
& & & MSE & MAE & MSE & MAE & MSE & MAE & MSE & MAE & MSE & MAE & MSE & MAE & MSE & MAE & MSE & MAE & MSE & MAE \\
\midrule

\multirow{20}{*}{Agriculture}
& \multirow{5}{*}{Uni-modal}
& 6   & 0.078 & 0.205 & 0.173 & 0.303 & 0.156 & 0.302 & 0.074 & 0.195 & 0.066 & 0.160 & 0.272 & 0.384 & 0.067 & 0.180 & 0.059 & 0.160 & 0.199 & 0.315 \\
& & 8   & 0.091 & 0.217 & 0.266 & 0.373 & 0.173 & 0.320 & 0.090 & 0.219 & 0.084 & 0.183 & 0.315 & 0.403 & 0.079 & 0.188 & 0.079 & 0.183 & 0.208 & 0.311 \\
& & 10  & 0.110 & 0.248 & 0.362 & 0.450 & 0.206 & 0.343 & 0.105 & 0.247 & 0.108 & 0.209 & 0.490 & 0.530 & 0.102 & 0.219 & 0.101 & 0.206 & 0.305 & 0.390 \\
& & 12  & 0.134 & 0.269 & 0.436 & 0.495 & 0.247 & 0.371 & 0.132 & 0.273 & 0.142 & 0.241 & 0.578 & 0.567 & 0.134 & 0.247 & 0.131 & 0.238 & 0.419 & 0.451 \\
& & Avg & 0.103 & 0.235 & 0.309 & 0.405 & 0.195 & 0.334 & 0.100 & 0.234 & 0.100 & 0.198 & 0.414 & 0.471 & 0.096 & 0.209 & 0.093 & 0.197 & 0.283 & 0.367 \\
\cmidrule(lr){2-21}

& \multirow{5}{*}{MM-TSFLib}
& 6   & 0.076 & 0.198 & 0.143 & 0.267 & 0.093 & 0.203 & 0.073 & 0.194 & 0.067 & 0.161 & 0.202 & 0.319 & 0.072 & 0.202 & 0.061 & 0.163 & 0.158 & 0.276 \\
\bottomrule
\end{tabular}
}
\end{table*}

\begin{table*}[t]
\centering
\tiny
\setlength{\tabcolsep}{1.6pt}
\renewcommand{\arraystretch}{0.92}
\caption{Detailed forecasting results on Health, Security, and Traffic under textual perturbation ratio $\rho=0.5$. Lower MSE/MAE indicates better performance.}
\label{tab:complete_perturbation_05_results_health_security_traffic}
\resizebox{\textwidth}{!}{
\begin{tabular}{llc|cc|cc|cc|cc|cc|cc|cc|cc|cc}
\toprule
\multirow{2}{*}{Dataset} & \multirow{2}{*}{Method} & \multirow{2}{*}{$T$}
& \multicolumn{2}{c|}{Autoformer}
& \multicolumn{2}{c|}{Crossformer}
& \multicolumn{2}{c|}{DLinear}
& \multicolumn{2}{c|}{FEDformer}
& \multicolumn{2}{c|}{FiLM}
& \multicolumn{2}{c|}{Informer}
& \multicolumn{2}{c|}{iTransformer}
& \multicolumn{2}{c|}{PatchTST}
& \multicolumn{2}{c}{Transformer} \\
\cmidrule(lr){4-5} \cmidrule(lr){6-7} \cmidrule(lr){8-9} \cmidrule(lr){10-11} \cmidrule(lr){12-13} \cmidrule(lr){14-15} \cmidrule(lr){16-17} \cmidrule(lr){18-19} \cmidrule(lr){20-21}
& & & MSE & MAE & MSE & MAE & MSE & MAE & MSE & MAE & MSE & MAE & MSE & MAE & MSE & MAE & MSE & MAE & MSE & MAE \\
\midrule

\multirow{20}{*}{Health}
& \multirow{5}{*}{Uni-modal}
& 12  & 1.512 & 0.919 & 1.216 & 0.748 & 1.579 & 0.855 & 1.131 & 0.787 & 1.244 & 0.749 & 1.250 & 0.765 & 1.061 & 0.626 & 0.962 & 0.624 & 1.358 & 0.803 \\
& & 24  & 1.875 & 1.025 & 1.434 & 0.798 & 1.656 & 0.879 & 1.549 & 0.964 & 1.525 & 0.832 & 1.642 & 0.819 & 1.811 & 0.784 & 1.578 & 0.758 & 1.393 & 0.761 \\
& & 36  & 1.971 & 1.040 & 1.450 & 0.781 & 1.692 & 0.892 & 1.622 & 0.982 & 1.692 & 0.891 & 1.534 & 0.824 & 2.187 & 0.885 & 1.908 & 0.857 & 1.555 & 0.793 \\
& & 48  & 1.881 & 0.987 & 1.591 & 0.834 & 1.731 & 0.913 & 1.647 & 0.980 & 1.790 & 0.931 & 1.662 & 0.865 & 2.240 & 0.928 & 2.131 & 0.911 & 1.533 & 0.835 \\
& & Avg & 1.810 & 0.993 & 1.423 & 0.790 & 1.664 & 0.885 & 1.487 & 0.928 & 1.563 & 0.851 & 1.522 & 0.818 & 1.825 & 0.806 & 1.645 & 0.788 & 1.460 & 0.798 \\
\cmidrule(lr){2-21}

& \multirow{5}{*}{MM-TSFLib}
& 12  & 1.449 & 0.885 & 1.165 & 0.719 & 1.483 & 0.841 & 1.190 & 0.805 & 1.186 & 0.734 & 1.120 & 0.727 & 0.999 & 0.624 & 0.879 & 0.603 & 1.171 & 0.762 \\
\bottomrule
\end{tabular}
}
\end{table*}

\begin{table*}[t]
\centering
\tiny
\setlength{\tabcolsep}{1.6pt}
\renewcommand{\arraystretch}{0.92}
\caption{Detailed forecasting results on Agriculture, Climate, and Economy under textual perturbation ratio $\rho=0.7$. Lower MSE/MAE indicates better performance.}
\label{tab:complete_perturbation_07_results_agriculture_climate_economy}
\resizebox{\textwidth}{!}{
\begin{tabular}{llc|cc|cc|cc|cc|cc|cc|cc|cc|cc}
\toprule
\multirow{2}{*}{Dataset} & \multirow{2}{*}{Method} & \multirow{2}{*}{$T$}
& \multicolumn{2}{c|}{Autoformer}
& \multicolumn{2}{c|}{Crossformer}
& \multicolumn{2}{c|}{DLinear}
& \multicolumn{2}{c|}{FEDformer}
& \multicolumn{2}{c|}{FiLM}
& \multicolumn{2}{c|}{Informer}
& \multicolumn{2}{c|}{iTransformer}
& \multicolumn{2}{c|}{PatchTST}
& \multicolumn{2}{c}{Transformer} \\
\cmidrule(lr){4-5} \cmidrule(lr){6-7} \cmidrule(lr){8-9} \cmidrule(lr){10-11} \cmidrule(lr){12-13} \cmidrule(lr){14-15} \cmidrule(lr){16-17} \cmidrule(lr){18-19} \cmidrule(lr){20-21}
& & & MSE & MAE & MSE & MAE & MSE & MAE & MSE & MAE & MSE & MAE & MSE & MAE & MSE & MAE & MSE & MAE & MSE & MAE \\
\midrule

\multirow{20}{*}{Agriculture}
& \multirow{5}{*}{Uni-modal}
& 6   & 0.078 & 0.205 & 0.173 & 0.303 & 0.156 & 0.302 & 0.074 & 0.195 & 0.066 & 0.160 & 0.272 & 0.384 & 0.067 & 0.180 & 0.059 & 0.160 & 0.199 & 0.315 \\
& & 8   & 0.091 & 0.217 & 0.266 & 0.373 & 0.173 & 0.320 & 0.090 & 0.219 & 0.084 & 0.183 & 0.315 & 0.403 & 0.079 & 0.188 & 0.079 & 0.183 & 0.208 & 0.311 \\
& & 10  & 0.110 & 0.248 & 0.362 & 0.450 & 0.206 & 0.343 & 0.105 & 0.247 & 0.108 & 0.209 & 0.490 & 0.530 & 0.102 & 0.219 & 0.101 & 0.206 & 0.305 & 0.390 \\
& & 12  & 0.134 & 0.269 & 0.436 & 0.495 & 0.247 & 0.371 & 0.132 & 0.273 & 0.142 & 0.241 & 0.578 & 0.567 & 0.134 & 0.247 & 0.131 & 0.238 & 0.419 & 0.451 \\
& & Avg & 0.103 & 0.235 & 0.309 & 0.405 & 0.195 & 0.334 & 0.100 & 0.234 & 0.100 & 0.198 & 0.414 & 0.471 & 0.096 & 0.209 & 0.093 & 0.197 & 0.283 & 0.367 \\
\cmidrule(lr){2-21}

& \multirow{5}{*}{MM-TSFLib}
& 6   & 0.076 & 0.198 & 0.138 & 0.259 & 0.103 & 0.227 & 0.073 & 0.194 & 0.067 & 0.161 & 0.205 & 0.322 & 0.071 & 0.198 & 0.061 & 0.163 & 0.159 & 0.277 \\
\bottomrule
\end{tabular}
}
\end{table*}

\begin{table*}[t]
\centering
\tiny
\setlength{\tabcolsep}{1.6pt}
\renewcommand{\arraystretch}{0.92}
\caption{Detailed forecasting results on Health, Security, and Traffic under textual perturbation ratio $\rho=0.7$. Lower MSE/MAE indicates better performance.}
\label{tab:complete_perturbation_07_results_health_security_traffic}
\resizebox{\textwidth}{!}{
\begin{tabular}{llc|cc|cc|cc|cc|cc|cc|cc|cc|cc}
\toprule
\multirow{2}{*}{Dataset} & \multirow{2}{*}{Method} & \multirow{2}{*}{$T$}
& \multicolumn{2}{c|}{Autoformer}
& \multicolumn{2}{c|}{Crossformer}
& \multicolumn{2}{c|}{DLinear}
& \multicolumn{2}{c|}{FEDformer}
& \multicolumn{2}{c|}{FiLM}
& \multicolumn{2}{c|}{Informer}
& \multicolumn{2}{c|}{iTransformer}
& \multicolumn{2}{c|}{PatchTST}
& \multicolumn{2}{c}{Transformer} \\
\cmidrule(lr){4-5} \cmidrule(lr){6-7} \cmidrule(lr){8-9} \cmidrule(lr){10-11} \cmidrule(lr){12-13} \cmidrule(lr){14-15} \cmidrule(lr){16-17} \cmidrule(lr){18-19} \cmidrule(lr){20-21}
& & & MSE & MAE & MSE & MAE & MSE & MAE & MSE & MAE & MSE & MAE & MSE & MAE & MSE & MAE & MSE & MAE & MSE & MAE \\
\midrule

\multirow{20}{*}{Health}
& \multirow{5}{*}{Uni-modal}
& 12  & 1.512 & 0.919 & 1.216 & 0.748 & 1.579 & 0.855 & 1.131 & 0.787 & 1.244 & 0.749 & 1.250 & 0.765 & 1.061 & 0.626 & 0.962 & 0.624 & 1.358 & 0.803 \\
& & 24  & 1.875 & 1.025 & 1.434 & 0.798 & 1.656 & 0.879 & 1.549 & 0.964 & 1.525 & 0.832 & 1.642 & 0.819 & 1.811 & 0.784 & 1.578 & 0.758 & 1.393 & 0.761 \\
& & 36  & 1.971 & 1.040 & 1.450 & 0.781 & 1.692 & 0.892 & 1.622 & 0.982 & 1.692 & 0.891 & 1.534 & 0.824 & 2.187 & 0.885 & 1.908 & 0.857 & 1.555 & 0.793 \\
& & 48  & 1.881 & 0.987 & 1.591 & 0.834 & 1.731 & 0.913 & 1.647 & 0.980 & 1.790 & 0.931 & 1.662 & 0.865 & 2.240 & 0.928 & 2.131 & 0.911 & 1.533 & 0.835 \\
& & Avg & 1.810 & 0.993 & 1.423 & 0.790 & 1.664 & 0.885 & 1.487 & 0.928 & 1.563 & 0.851 & 1.522 & 0.818 & 1.825 & 0.806 & 1.645 & 0.788 & 1.460 & 0.798 \\
\cmidrule(lr){2-21}

& \multirow{5}{*}{MM-TSFLib}
& 12  & 1.485 & 0.894 & 1.175 & 0.724 & 1.483 & 0.841 & 1.155 & 0.790 & 1.185 & 0.729 & 1.140 & 0.722 & 0.990 & 0.621 & 0.891 & 0.608 & 1.177 & 0.768 \\
\bottomrule
\end{tabular}
}
\end{table*}

\begin{table*}[t]
\centering
\tiny
\setlength{\tabcolsep}{1.6pt}
\renewcommand{\arraystretch}{0.92}
\caption{Detailed forecasting results on Agriculture, Climate, and Economy under textual perturbation ratio $\rho=0.9$. Lower MSE/MAE indicates better performance.}
\label{tab:complete_perturbation_09_results_agriculture_climate_economy}
\resizebox{\textwidth}{!}{
\begin{tabular}{llc|cc|cc|cc|cc|cc|cc|cc|cc|cc}
\toprule
\multirow{2}{*}{Dataset} & \multirow{2}{*}{Method} & \multirow{2}{*}{$T$}
& \multicolumn{2}{c|}{Autoformer}
& \multicolumn{2}{c|}{Crossformer}
& \multicolumn{2}{c|}{DLinear}
& \multicolumn{2}{c|}{FEDformer}
& \multicolumn{2}{c|}{FiLM}
& \multicolumn{2}{c|}{Informer}
& \multicolumn{2}{c|}{iTransformer}
& \multicolumn{2}{c|}{PatchTST}
& \multicolumn{2}{c}{Transformer} \\
\cmidrule(lr){4-5} \cmidrule(lr){6-7} \cmidrule(lr){8-9} \cmidrule(lr){10-11} \cmidrule(lr){12-13} \cmidrule(lr){14-15} \cmidrule(lr){16-17} \cmidrule(lr){18-19} \cmidrule(lr){20-21}
& & & MSE & MAE & MSE & MAE & MSE & MAE & MSE & MAE & MSE & MAE & MSE & MAE & MSE & MAE & MSE & MAE & MSE & MAE \\
\midrule

\multirow{20}{*}{Agriculture}
& \multirow{5}{*}{Uni-modal}
& 6   & 0.078 & 0.205 & 0.173 & 0.303 & 0.156 & 0.302 & 0.074 & 0.195 & 0.066 & 0.160 & 0.272 & 0.384 & 0.067 & 0.180 & 0.059 & 0.160 & 0.199 & 0.315 \\
& & 8   & 0.091 & 0.217 & 0.266 & 0.373 & 0.173 & 0.320 & 0.090 & 0.219 & 0.084 & 0.183 & 0.315 & 0.403 & 0.079 & 0.188 & 0.079 & 0.183 & 0.208 & 0.311 \\
& & 10  & 0.110 & 0.248 & 0.362 & 0.450 & 0.206 & 0.343 & 0.105 & 0.247 & 0.108 & 0.209 & 0.490 & 0.530 & 0.102 & 0.219 & 0.101 & 0.206 & 0.305 & 0.390 \\
& & 12  & 0.134 & 0.269 & 0.436 & 0.495 & 0.247 & 0.371 & 0.132 & 0.273 & 0.142 & 0.241 & 0.578 & 0.567 & 0.134 & 0.247 & 0.131 & 0.238 & 0.419 & 0.451 \\
& & Avg & 0.103 & 0.235 & 0.309 & 0.405 & 0.195 & 0.334 & 0.100 & 0.234 & 0.100 & 0.198 & 0.414 & 0.471 & 0.096 & 0.209 & 0.093 & 0.197 & 0.283 & 0.367 \\
\cmidrule(lr){2-21}

& \multirow{5}{*}{MM-TSFLib}
& 6   & 0.076 & 0.199 & 0.139 & 0.260 & 0.095 & 0.209 & 0.073 & 0.194 & 0.067 & 0.161 & 0.204 & 0.320 & 0.070 & 0.199 & 0.061 & 0.163 & 0.159 & 0.277 \\
\bottomrule
\end{tabular}
}
\end{table*}

\begin{table*}[t]
\centering
\tiny
\setlength{\tabcolsep}{1.6pt}
\renewcommand{\arraystretch}{0.92}
\caption{Detailed forecasting results on Health, Security, and Traffic under textual perturbation ratio $\rho=0.9$. Lower MSE/MAE indicates better performance.}
\label{tab:complete_perturbation_09_results_health_security_traffic}
\resizebox{\textwidth}{!}{
\begin{tabular}{llc|cc|cc|cc|cc|cc|cc|cc|cc|cc}
\toprule
\multirow{2}{*}{Dataset} & \multirow{2}{*}{Method} & \multirow{2}{*}{$T$}
& \multicolumn{2}{c|}{Autoformer}
& \multicolumn{2}{c|}{Crossformer}
& \multicolumn{2}{c|}{DLinear}
& \multicolumn{2}{c|}{FEDformer}
& \multicolumn{2}{c|}{FiLM}
& \multicolumn{2}{c|}{Informer}
& \multicolumn{2}{c|}{iTransformer}
& \multicolumn{2}{c|}{PatchTST}
& \multicolumn{2}{c}{Transformer} \\
\cmidrule(lr){4-5} \cmidrule(lr){6-7} \cmidrule(lr){8-9} \cmidrule(lr){10-11} \cmidrule(lr){12-13} \cmidrule(lr){14-15} \cmidrule(lr){16-17} \cmidrule(lr){18-19} \cmidrule(lr){20-21}
& & & MSE & MAE & MSE & MAE & MSE & MAE & MSE & MAE & MSE & MAE & MSE & MAE & MSE & MAE & MSE & MAE & MSE & MAE \\
\midrule

\multirow{20}{*}{Health}
& \multirow{5}{*}{Uni-modal}
& 12  & 1.512 & 0.919 & 1.216 & 0.748 & 1.579 & 0.855 & 1.131 & 0.787 & 1.244 & 0.749 & 1.250 & 0.765 & 1.061 & 0.626 & 0.962 & 0.624 & 1.358 & 0.803 \\
& & 24  & 1.875 & 1.025 & 1.434 & 0.798 & 1.656 & 0.879 & 1.549 & 0.964 & 1.525 & 0.832 & 1.642 & 0.819 & 1.811 & 0.784 & 1.578 & 0.758 & 1.393 & 0.761 \\
& & 36  & 1.971 & 1.040 & 1.450 & 0.781 & 1.692 & 0.892 & 1.622 & 0.982 & 1.692 & 0.891 & 1.534 & 0.824 & 2.187 & 0.885 & 1.908 & 0.857 & 1.555 & 0.793 \\
& & 48  & 1.881 & 0.987 & 1.591 & 0.834 & 1.731 & 0.913 & 1.647 & 0.980 & 1.790 & 0.931 & 1.662 & 0.865 & 2.240 & 0.928 & 2.131 & 0.911 & 1.533 & 0.835 \\
& & Avg & 1.810 & 0.993 & 1.423 & 0.790 & 1.664 & 0.885 & 1.487 & 0.928 & 1.563 & 0.851 & 1.522 & 0.818 & 1.825 & 0.806 & 1.645 & 0.788 & 1.460 & 0.798 \\
\cmidrule(lr){2-21}

& \multirow{5}{*}{MM-TSFLib}
& 12  & 1.523 & 0.918 & 1.170 & 0.720 & 1.484 & 0.841 & 1.171 & 0.795 & 1.186 & 0.733 & 1.085 & 0.712 & 1.006 & 0.624 & 0.907 & 0.614 & 1.179 & 0.767 \\
\bottomrule
\end{tabular}
}
\end{table*}

\clearpage
\section{Broader Impacts}
\label{app:broader_impacts}

This work studies robust multimodal time-series forecasting under noisy textual conditions. The proposed method may be useful for applications where numerical time-series data are accompanied by textual information, such as event descriptions, reports, or news summaries. By improving the model's ability to handle corrupted or incomplete textual inputs, this work may help reduce the sensitivity of forecasting systems to imperfect auxiliary text.

However, the practical impact of the method depends on the application scenario and the quality of the available data. Forecasting errors may still occur, especially when textual information is severely corrupted, biased, or inconsistent with the numerical signals. In high-stakes domains such as healthcare, energy management, public safety, or transportation, inaccurate predictions may lead to inappropriate decisions if they are used without sufficient human verification. Therefore, the proposed method should not be viewed as a standalone decision-making system. Practical deployment should involve careful validation, uncertainty estimation, and supervision from domain experts.
%%%%%%%%%%%%%%%%%%%%%%%%%%%%%%%%%%%%%%%%%%%%%%%%%%%%%%%%%%%%

% \newpage
% \input{checklist.tex}
\clearpage
\input{checklist.tex}

\end{document}

%% file: checklist.tex
\section*{NeurIPS Paper Checklist}

\begin{enumerate}

% \item {\bf Claims}
%     \item[] Question: Do the main claims made in the abstract and introduction accurately reflect the paper's contributions and scope?
%     \item[] Answer: \answerTODO{} % Replace by \answerYes{}, \answerNo{}, or \answerNA{}.
%     \item[] Justification: \justificationTODO{}
%     \item[] Guidelines:
%     \begin{itemize}
%         \item The answer \answerNA{} means that the abstract and introduction do not include the claims made in the paper.
%         \item The abstract and/or introduction should clearly state the claims made, including the contributions made in the paper and important assumptions and limitations. A \answerNo{} or \answerNA{} answer to this question will not be perceived well by the reviewers. 
%         \item The claims made should match theoretical and experimental results, and reflect how much the results can be expected to generalize to other settings. 
%         \item It is fine to include aspirational goals as motivation as long as it is clear that these goals are not attained by the paper. 
%     \end{itemize}

\item {\bf Claims}
    \item[] Question: Do the main claims made in the abstract and introduction accurately reflect the paper's contributions and scope?
    \item[] Answer: \answerYes{}
    \item[] Justification: The abstract and introduction clearly state the main contributions of our work, including the proposed method, its motivation, and the forecasting problem it addresses. The claims are supported by the experimental results and ablation studies reported in the experiments section, and we avoid making claims beyond the evaluated datasets and settings.

\item {\bf Limitations}
    \item[] Question: Does the paper discuss the limitations of the work performed by the authors?
    \item[] Answer: \answerYes{}
    \item[] Justification: The paper discusses potential limitations of the proposed method, including cases where the results are less satisfactory on certain datasets or backbones, and provides possible reasons for these observations.

\item {\bf Theory assumptions and proofs}
    \item[] Question: For each theoretical result, does the paper provide the full set of assumptions and a complete (and correct) proof?
    \item[] Answer: \answerYes{}
    \item[] Justification: The paper provides the full set of assumptions for the theoretical results and includes detailed proofs in the appendix. The theoretical derivations are presented with corresponding explanations to support the validity of the proposed method.

\item {\bf Experimental result reproducibility}
    \item[] Question: Does the paper fully disclose all the information needed to reproduce the main experimental results of the paper to the extent that it affects the main claims and/or conclusions of the paper (regardless of whether the code and data are provided or not)?
    \item[] Answer: \answerYes{}
    \item[] Justification: The paper provides the necessary details for reproducing the main experimental results, including the experimental settings, evaluation metrics, hyperparameter configurations, backbone models, perturbation settings, and random seed setup. These details support the reproducibility of the reported results and the main empirical conclusions.

\item {\bf Open access to data and code}
    \item[] Question: Does the paper provide open access to the data and code, with sufficient instructions to faithfully reproduce the main experimental results, as described in supplemental material?
    \item[] Answer: \answerYes{}
    \item[] Justification: The code and datasets are provided in the supplemental material, together with instructions for reproducing the main experimental results. 

\item {\bf Experimental setting/details}
    \item[] Question: Does the paper specify all the training and test details (e.g., data splits, hyperparameters, how they were chosen, type of optimizer) necessary to understand the results?
    \item[] Answer: \answerYes{}
    \item[] Justification: The paper reports the necessary training and testing details, including data splits, hyperparameters, evaluation metrics, backbone configurations, perturbation settings, and random seed setup. Further implementation details are provided in the appendix and supplemental material.

\item {\bf Experiment statistical significance}
    \item[] Question: Does the paper report error bars suitably and correctly defined or other appropriate information about the statistical significance of the experiments?
    \item[] Answer: \answerNo{}
    \item[] Justification: The experiments are repeated with multiple random seeds, and the reported results are averaged across these runs. However, standard deviations, confidence intervals, or statistical significance tests are not explicitly reported due to space limitations.

\item {\bf Experiments compute resources}
    \item[] Question: For each experiment, does the paper provide sufficient information on the computer resources (type of compute workers, memory, time of execution) needed to reproduce the experiments?
    \item[] Answer: \answerYes{}
    \item[] Justification: The paper provides information on the computational resources used for the experiments, including the hardware environment, GPU configuration, memory requirements. These details help readers understand the computational cost of reproducing the reported experiments.
    
\item {\bf Code of ethics}
    \item[] Question: Does the research conducted in the paper conform, in every respect, with the NeurIPS Code of Ethics \url{https://neurips.cc/public/EthicsGuidelines}?
    \item[] Answer: \answerYes{}
    \item[] Justification: The research conducted in this paper conforms to the NeurIPS Code of Ethics. The work is based on benchmark textual-numerical time-series datasets and does not involve human subjects, sensitive personal information, or harmful data collection. The experiments and analyses are conducted in a manner that preserves research integrity and anonymity during submission.

\item {\bf Broader impacts}
    \item[] Question: Does the paper discuss both potential positive societal impacts and negative societal impacts of the work performed?
    \item[] Answer: \answerYes{}
    \item[] Justification: The paper discusses the broader impacts of the proposed method in Appendix. It describes the potential usefulness of robust multimodal time-series forecasting under noisy textual conditions, while also acknowledging risks related to inaccurate predictions, data quality, and over-reliance in high-stakes applications.
    
\item {\bf Safeguards}
    \item[] Question: Does the paper describe safeguards that have been put in place for responsible release of data or models that have a high risk for misuse (e.g., pre-trained language models, image generators, or scraped datasets)?
    \item[] Answer: \answerNA{}
    \item[] Justification: The paper does not release data or models that pose a high risk for misuse, such as generative models, scraped sensitive datasets, or systems intended for dual-use applications. The released materials are limited to datasets and code for multimodal time-series forecasting experiments, and the work does not involve high-risk model deployment or unsafe data release.

\item {\bf Licenses for existing assets}
    \item[] Question: Are the creators or original owners of assets (e.g., code, data, models), used in the paper, properly credited and are the license and terms of use explicitly mentioned and properly respected?
    \item[] Answer: \answerYes{}
    \item[] Justification: The paper properly credits the creators of the existing assets used in this work, including benchmark datasets, forecasting backbones, and pre-trained language models. Relevant papers and sources are cited, and the licenses and terms of use of these assets are respected. Details on the datasets and implementation resources are provided in the paper, appendix, or supplemental material.

\item {\bf New assets}
    \item[] Question: Are new assets introduced in the paper well documented and is the documentation provided alongside the assets?
    \item[] Answer: \answerYes{}
    \item[] Justification: The new assets introduced in this work, including the implementation of the proposed PA-RNet framework and related experimental scripts, are documented and provided in the supplemental material. The released materials are anonymized for submission when applicable.

\item {\bf Crowdsourcing and research with human subjects}
    \item[] Question: For crowdsourcing experiments and research with human subjects, does the paper include the full text of instructions given to participants and screenshots, if applicable, as well as details about compensation (if any)? 
    \item[] Answer: \answerNA{}
    \item[] Justification: The paper does not involve crowdsourcing experiments or research with human subjects. Therefore, participant instructions, screenshots, and compensation details are not applicable.

\item {\bf Institutional review board (IRB) approvals or equivalent for research with human subjects}
    \item[] Question: Does the paper describe potential risks incurred by study participants, whether such risks were disclosed to the subjects, and whether Institutional Review Board (IRB) approvals (or an equivalent approval/review based on the requirements of your country or institution) were obtained?
    \item[] Answer: \answerNA{}
    \item[] Justification: This work does not involve human subjects or crowdsourcing experiments, so IRB approval or equivalent review is not applicable.

\item {\bf Declaration of LLM usage}
    \item[] Question: Does the paper describe the usage of LLMs if it is an important, original, or non-standard component of the core methods in this research? Note that if the LLM is used only for writing, editing, or formatting purposes and does \emph{not} impact the core methodology, scientific rigor, or originality of the research, declaration is not required.
    %this research? 
    \item[] Answer: \answerYes{}
    \item[] Justification: The paper clearly describes the use of a GPT2-based encoder as part of the proposed multimodal forecasting framework. Specifically, the encoder is used to process textual inputs and extract textual representations that are further integrated with numerical time-series features. Since the LLM component is part of the core methodology rather than being used only for writing, editing, or formatting, its usage is explicitly declared and discussed in the paper.

\end{enumerate}